\documentclass[journal]{IEEEtran}
\ifCLASSINFOpdf
\else
\fi
\usepackage{mathtools}
\usepackage{color}
\usepackage{algorithm}
\usepackage[noend]{algpseudocode}

\usepackage{times}
\usepackage{epsfig}
\usepackage{graphicx}
\usepackage{amssymb}
\usepackage{romannum}
\usepackage{color}

\usepackage{amsmath}
\usepackage{breqn}
\usepackage{multirow}
\usepackage{subfigure}
\usepackage{xurl}
\usepackage{caption}

\begin{document}
%
\title{Data Augmentation via Diffusion Model to Enhance AI Fairness}
%
%
%

 \author{Christina Hastings Blow,~Lijun Qian,~ Camille Gibson, ~Pamela Obiomon,~Xishuang Dong
\thanks{C. Hastings, L. Qian, P. Obiomon and X. Dong are with the Department of Electrical and Computer Engineering, Prairie View A\&M University, Texas A\&M University System, Prairie View, TX 77446, USA. C. Gibson is with the College of Juvenile Justice,
Executive Director of Texas Juvenile Crime Prevention Center, Prairie View A\&M University, Texas A\&M University System, Prairie View, TX 77446, USA. Email: chastings@pvamu.edu, liqian@pvamu.edu, cbgibson@pvamu.edu, phobiomon@pvamu.edu, xidong@pvamu.edu}
}

\maketitle

\begin{abstract}
AI fairness seeks to improve the transparency and explainability of AI systems by ensuring that their outcomes genuinely reflect the best interests of users. Data augmentation, which involves generating synthetic data from existing datasets, has gained significant attention as a solution to data scarcity. In particular, diffusion models have become a powerful technique for generating synthetic data, especially in fields like computer vision. This paper explores the potential of diffusion models to generate synthetic tabular data to improve AI fairness. The Tabular Denoising Diffusion Probabilistic Model (Tab-DDPM), a diffusion model adaptable to any tabular dataset and capable of handling various feature types, was utilized with different amounts of generated data for data augmentation. Additionally, reweighting samples from AIF360 was employed to further enhance AI fairness. Five traditional machine learning models—Decision Tree (DT), Gaussian Naive Bayes (GNB), K-Nearest Neighbors (KNN), Logistic Regression (LR), and Random Forest (RF)—were used to validate the proposed approach. Experimental results demonstrate that the synthetic data generated by Tab-DDPM improves fairness in binary classification.

\end{abstract}

\begin{IEEEkeywords}
AI Fairness, Diffusion Model, Data Scarcity,  AIF360
\end{IEEEkeywords}

%
\IEEEpeerreviewmaketitle

\section{Introduction }
\label{sec1}


In our rapidly evolving society, artificial intelligence (AI) has become a ubiquitous presence, influencing everyday activities like online banking and digital assistants. However, how can we ensure the fairness of AI-generated outcomes? AI fairness seeks to enhance the transparency and explainability of AI systems~\cite{li2023trustworthy}. It scrutinizes the results to determine if they genuinely consider the users' best interests. Additionally, guidelines are being established to ensure the safety of both corporations and consumers. Various fairness tools have been developed to address the growing need to mitigate AI biases~\cite{richardson2021framework}. For example, AIF360~\cite{bellamy2019ai} offers a comprehensive set of fairness metrics for datasets and models, explanations for these metrics, and algorithms to reduce bias in datasets and models concerning protected attributes such as sex and race.

Data augmentation~\cite{ding2019case} aims to generate synthetic data from existing datasets to enlarge the training data to enhance the machine learning performance~\cite{bansal2022systematic}. This technique increases both the quantity and variety of data available for training and testing models, eliminating the need for new data collection. Data augmentation can be achieved by either learning a generator, such as through GAN networks~\cite{alqahtani2021applications}, to create data from scratch, or by learning a set of transformations to apply to existing training set samples~\cite{cubuk2019autoaugment} . Both approaches enhance the performance of deep learning models by providing a more diverse and abundant dataset.

In recent years, diffusion models~\cite{yang2023diffusion} have emerged as a powerful technique for generating synthetic data to address data scarcity. For example, Villaiz{\'a}n-Vallelado~\textit{et. al} proposed a diffusion model for generating synthetic tabular data with three key enhancements: a conditioning attention mechanism, an encoder-decoder transformer as the denoising network, and dynamic masking~\cite{villaizan2024diffusion}. Nguyen~\textit{et. al} introduced a novel method for generating pixel-level semantic segmentation labels using the text-to-image generative model Stable Diffusion (SD)~\cite{nguyen2024dataset}, which incorporates uncertainty regions into the segmentation to account for imperfections in the pseudo-labels. Additionally, Hu~\textit{et. al} developed a novel diffusion GNN model called Syngand, capable of generating ligand and pharmacokinetic data end-to-end, providing a methodology for sampling pharmacokinetic data for existing ligands using this model~\cite{hu2024synthetic}.

This paper aims to investigate whether diffusion models can generate synthetic data to enhance AI fairness as well as machine learning performance. Tabular Denoising Diffusion Probabilistic Model (Tab-DDPM)~\cite{kotelnikov2023tabddpm} is a diffusion model that can be universally applied to any tabular dataset, handling all feature types. It uses multinomial diffusion for categorical and binary features, and Gaussian diffusion for numerical ones. Tab-DDPM effectively manages mixed data types and consistently generates high-quality synthetic data. It is used to conduct different increments of generated data samples, specifically 20,000, 100,000, and 150,000 samples. To further mitigate bias, reweighting samples was employed to recalibrate the data. This involves adjusting the significance or contribution of individual samples within the training dataset, making it possible to remove discrimination concerning sensitive attributes without altering existing labels~\cite{calders2009building}. We used techniques from AIF360~\cite{bellamy2019ai}  to determine these weights, based on the frequency counts associated with the sensitive attribute. To validate the proposed method, five traditional machine learning models were applied: Decision Tree (DT), Gaussian Naive Bayes (GNB), K Nearest Neighbor (KNN), Logistic Regression (LR), and Random Forest (RF). Experimental results indicate that the synthetic data generated by Tab-DDPM enhances the fairness of binary classification. For instance, both RF performance in binary classification and fairness evaluated by five evaluation metrics has been improved when enlarging the training data with the generated data.

The contributions of this paper can be summarized as:
\begin{enumerate}
\item Introduction of generative AI techniques for generating synthetic data to enhance AI fairness as well as machine learning performance.

\item Extensive experiments demonstrating that the fairness of different machine learning models can be improved with respect to various protected attributes.
\end{enumerate}

\section{Methodology}
\label{sec2}

This paper aims to examine the effectiveness of Tab-DDPM and sample reweighting in enhancing the fairness of traditional machine learning algorithms on classification tasks, focusing on two key AI techniques: diffusion models and sample reweighting.

\subsection{Tab-DDPM} 

Tab-DDPM~\cite{kotelnikov2023tabddpm} is a generative model for tabular data, an area of active research. Tabular datasets are often limited in size due to privacy concerns during data collection. Generative AI, like Tab-DDPM, can create new synthetic data without these privacy issues. It is a newly developed model capable of effectively generating new data from tabular datasets. In detail, the DDPM process consists of three main components: the forward process, the backward process, and the sampling procedure~\cite{chang2023design}. The forward process adds noise to the training data. The reverse process trains denoising networks to iteratively remove noise, differing from generative adversarial networks (GANs) by removing noise over two timesteps instead of one. The sampling procedure uses the optimized denoising network to generate novel data. It uses a Gaussian diffusion model for numerical data and a Multinomial diffusion model for categorical and binary features. Hyperparameters play a crucial role in Tab-DDPM, significantly influencing the model's effectiveness. The general framework of Tab-DDPM is shown as Figure~\ref{Fig_tabddpm} below.

\begin{figure} [ht]
 	\centering
	\includegraphics[width=1.\linewidth]{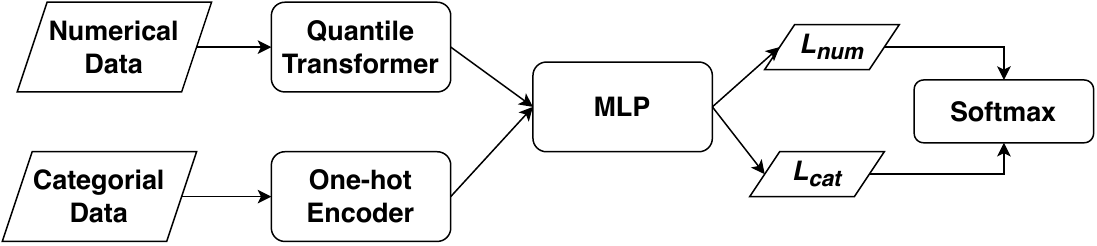}
	\caption{Tab-DDPM framework.}
	\label{Fig_tabddpm}
\end{figure}

The numerical and categorical data were represented through two branches: quantile transformer for numerical data and one-hot encoding for categorical data. These new data representations were then fed into a DDPM process utilizing multilayer perceptrons (MLP) to minimize two types of losses $L_{num}$ and $L_{cat}$ using softmax function.

\subsection{Reweighting Samples}

Reweighting samples is a preprocessing technique that adjusts the significance or contribution of samples within a training dataset. Weights are strategically assigned making it possible to render datasets free from discrimination pertaining to sensitive attributes without altering existing labels. One such approach is by based on the frequency counts associated with the sensitive attribute~\cite{calders2009building}.

This paper utilized the reweighting sample technique from the AIF360 toolbox for reweighting during the preprocessing phase. The contribution of the reweighting process comprises the training dataset with generated data of different increments with these samples containing attributes (including a sensitive attribute) and labels along with the specification of the sensitive attribute. The result being a transformed dataset where sample weights are adjusted for the sensitive attributes, mitigating potential classification bias. Throughout the reweighting process, an analysis of the allocation of the sensitive attributes within various groups is conducted. This analysis informs the calculation of reweighting coefficients, which, in turn, amends the sample weights to encourage a more uniform distribution across groups \cite{blow2024comprehensive}. For instance, given a sensitive (protected) attribute, the privileged group of these samples includes the samples with the positive sensitive attribute while  the unprivileged group of samples includes the samples with the negative sensitive attribute. 

\subsection{Proposed Method}

\begin{figure} [ht]
 	\centering
	\includegraphics[width=1.\linewidth]{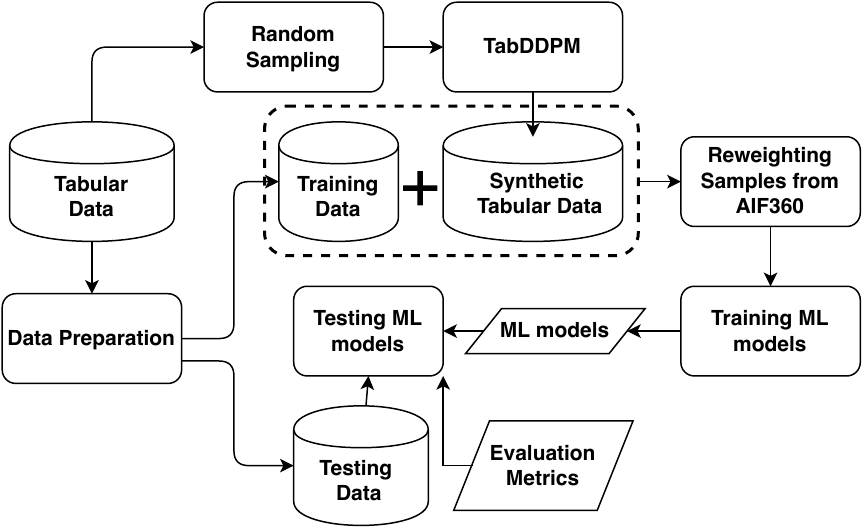}
	\caption{Flow of the proposed method.}
	\label{Fig_model}
\end{figure}

The flow of the proposed method is depicted in Figure~\ref{Fig_model}. The process begins with the random sampling of data, which serves as input to Tab-DDPM to generate synthetic tabular data. This synthetic data is then combined with the original training data to create a comprehensive dataset for training the ML model. In addition, reweighting samples from AIF360 is employ to adjust weights of different categories of samples to enhance fairness. Finally, the trained ML model is evaluated using test data, with performance assessed through multiple evaluation metrics, including various fairness metrics.

TabDDPM processes numerical and categorical features using Gaussian diffusion and Multinomial diffusion, respectively.  For instance, a tabular data sample $x =~<x_{num_{1}} , ... , x_{num_{N}}, x_{c_{1}}, ... , x_{c_{N}}>$ consists of $num_{N}$ numerical features and $c_{N}$ categorical features. Specifically, TabDDPM applies a Gaussian quantile transformation to process each categorical feature through a separate forward diffusion process, where noise components for all features are sampled independently. The reverse diffusion step in TabDDPM is executed by a multi-layer neural network, which produces an output with the same dimensionality as the input.

Reweighting samples involves adjusting the weights of four categories: $w_{pp}$ (weight of positive privileged samples), $w_{pup}$ (weight of positive unprivileged samples), $w_{np}$ (weight of negative privileged samples), and $w_{nup}$ (weight of negative unprivileged samples), as outlined below."

\begin{equation}
w_{pp} = \frac{N_{p}}{N_{total}} \times \frac{N_{pos}}{N_{pp}} 
\end{equation}

\begin{equation}
w_{pup} = \frac{N_{up}}{N_{total}} \times \frac{N_{pos}}{N_{pup}} 
\end{equation}

\begin{equation}
w_{np} = \frac{N_{p}}{N_{total}} \times \frac{N_{neg}}{N_{np}} 
\end{equation}

\begin{equation}
w_{nup} = \frac{N_{up}}{N_{total}} \times \frac{N_{neg}}{N_{up}} 
\end{equation}

where 

$N_{p}$: the number of samples in the privileged group.

$N_{pp}$:  the number of samples with the positive class in the privileged group. 

$N_{np}$:  the number of samples with the negative class in the privileged group. 

$N_{up}$:   the number of samples in the unprivileged group.

$N_{pup}$:  the number of samples with the positive class in the unprivileged group. 

$N_{nup}$: the number of samples with the negative class in the unprivileged group. 

$N_{pos}$: the number of samples with the positive class.

$N_{neg}$:  the number of samples with the negative class.

$N_{total}$: the number of samples.

\section{Experiment }
\label{sec4}

\subsection{Datasets}

This study utilized the Adult Income dataset and applied Tab-DDPM to generate new synthetic data, aiming to assess the combined effectiveness of data augmentation and sample reweighting in mitigating fairness issues.

\textbf{Adult Income Dataset:} The dataset consists of $48,842$ samples with $14$ attributes, designed to predict whether an individual's income exceeds \$50K/year based on census data~\cite{misc_adult_2}. It was divided into training ($28,048$ samples), testing ($16,281$ samples), and validation ($6,513$ samples) sets. The attributes were categorized into eight categorical and six numerical features.

\begin{figure*}

\begin{tabular}{cccc}

 \includegraphics[width=0.2\textwidth]{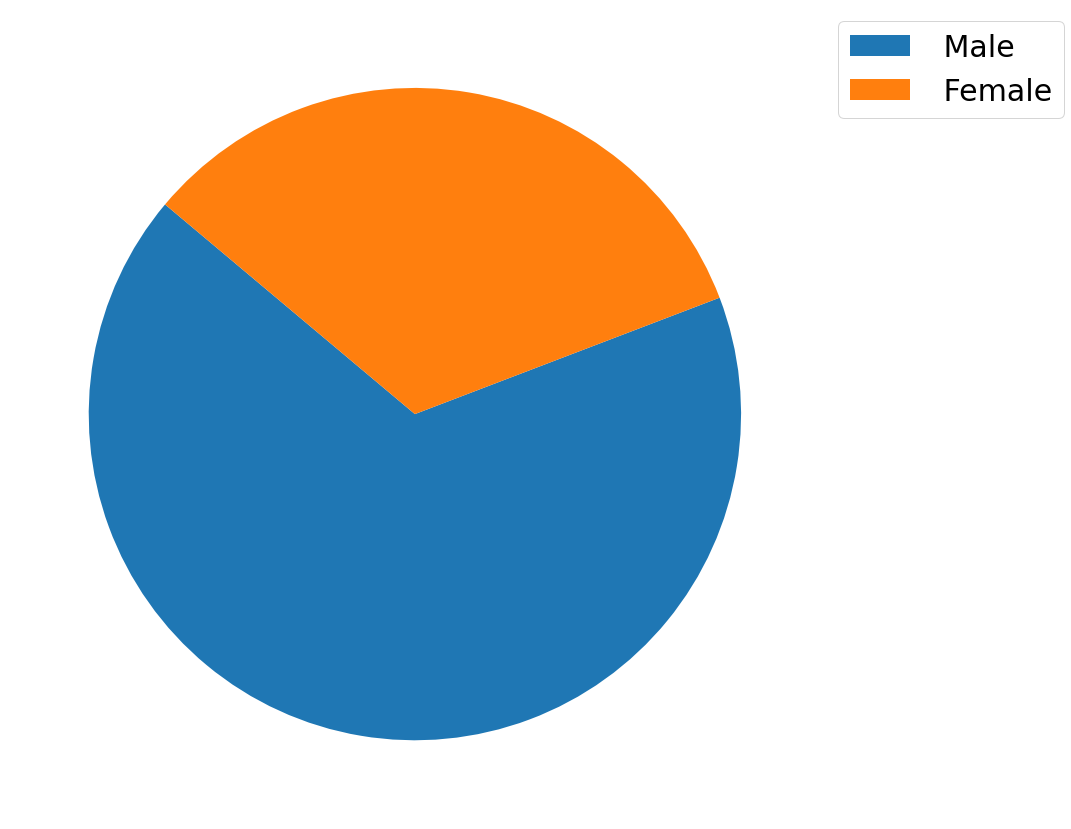} &   \includegraphics[width=0.2\textwidth]{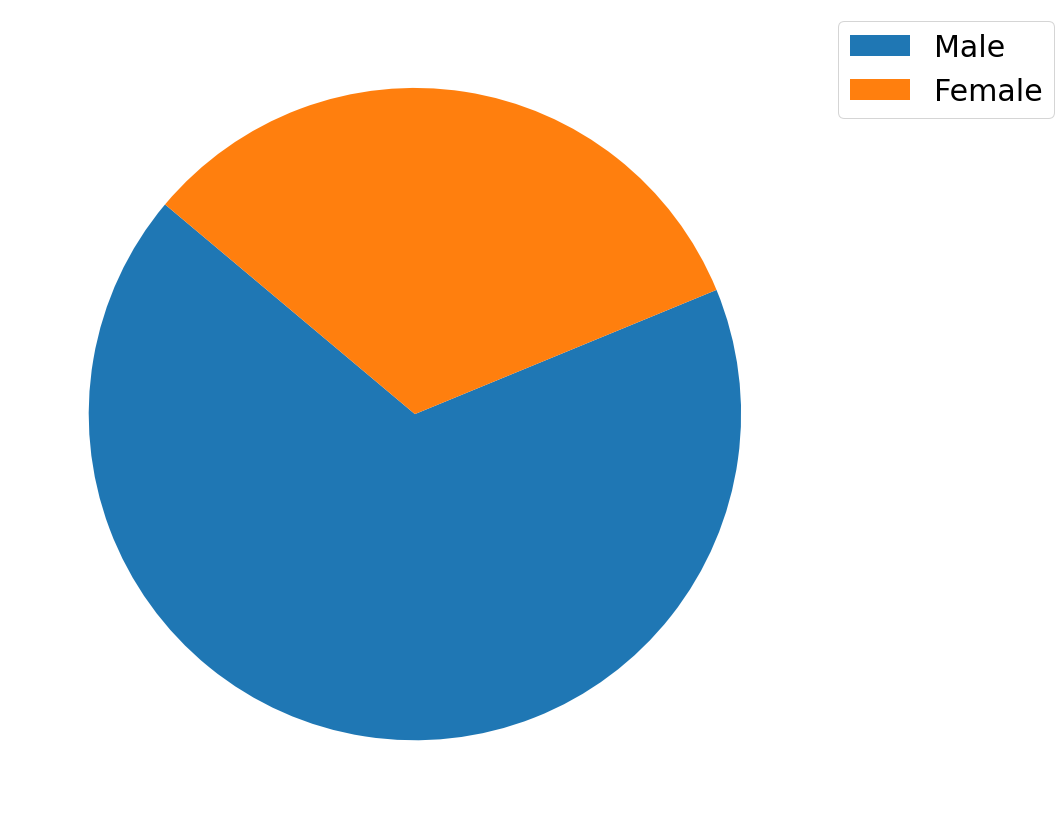} &  \includegraphics[width=0.2\textwidth]{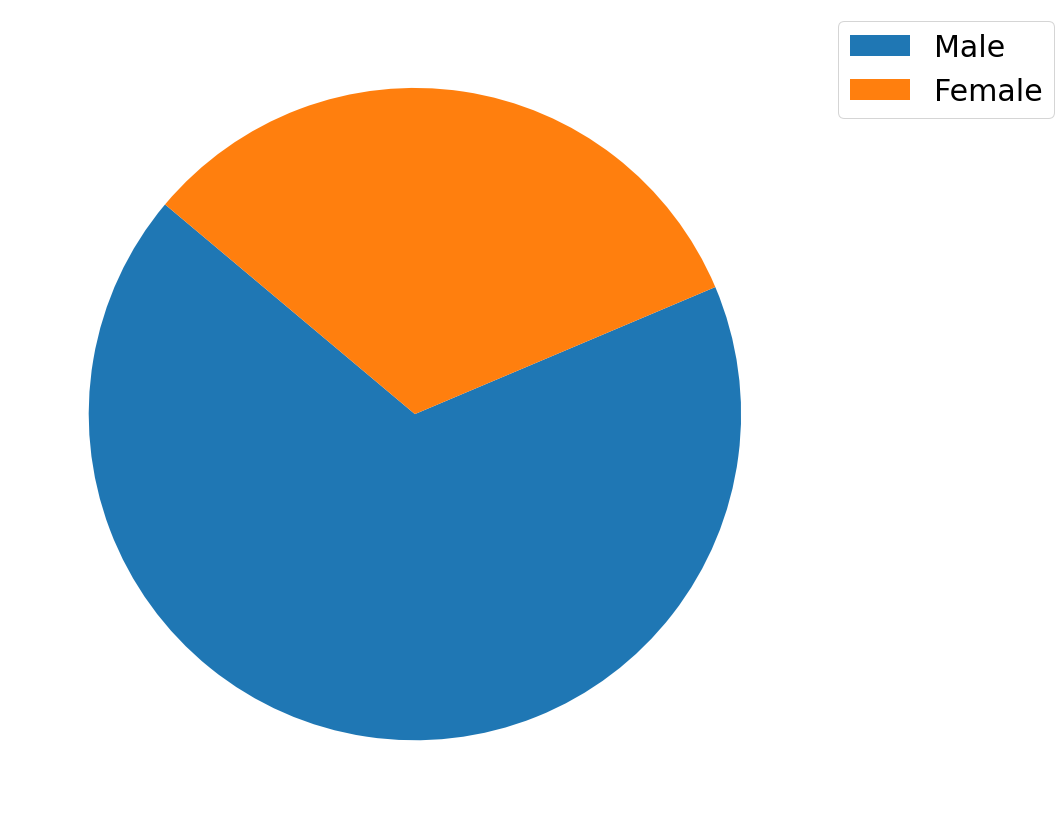} &  \includegraphics[width=0.2\textwidth]{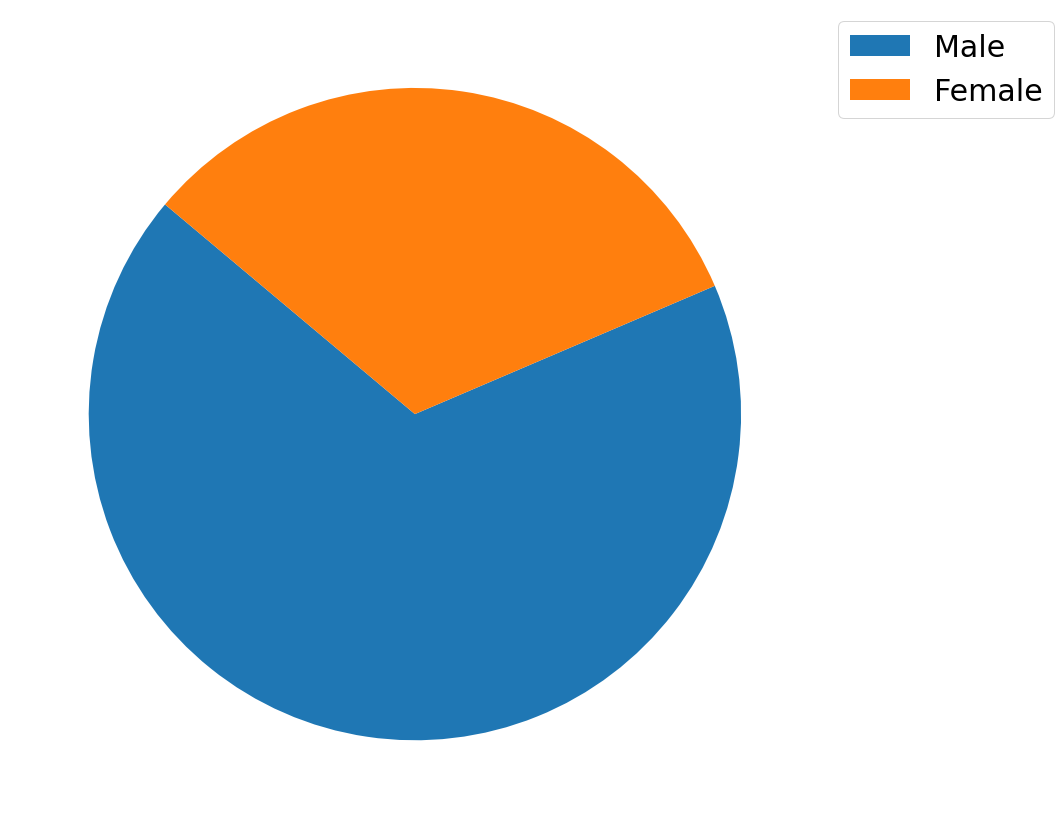}  \\
  (1) Original sex &   (2) 20,000 sex case  &   (3) 100,000 sex case &   (4) 150,000 sex case \\
  
  \includegraphics[width=0.24\textwidth]{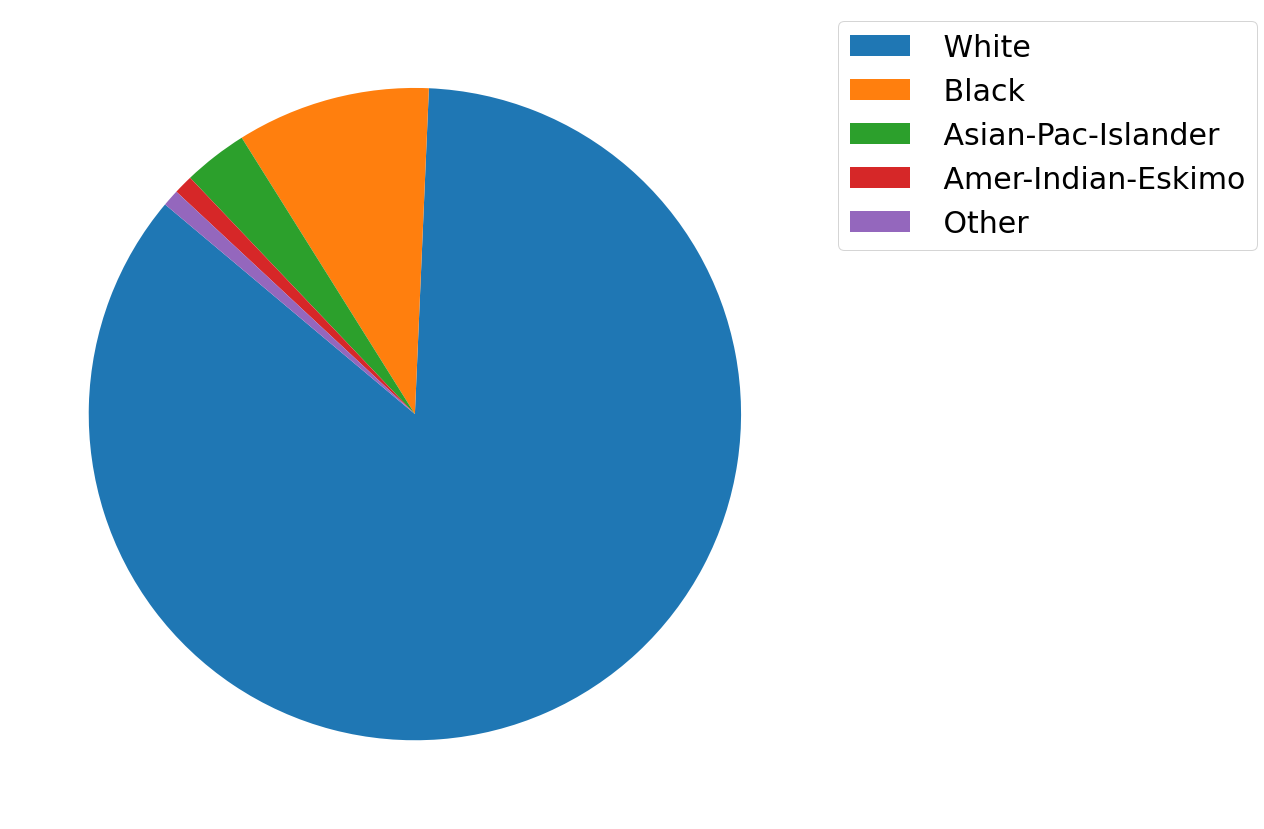} &   \includegraphics[width=0.24\textwidth]{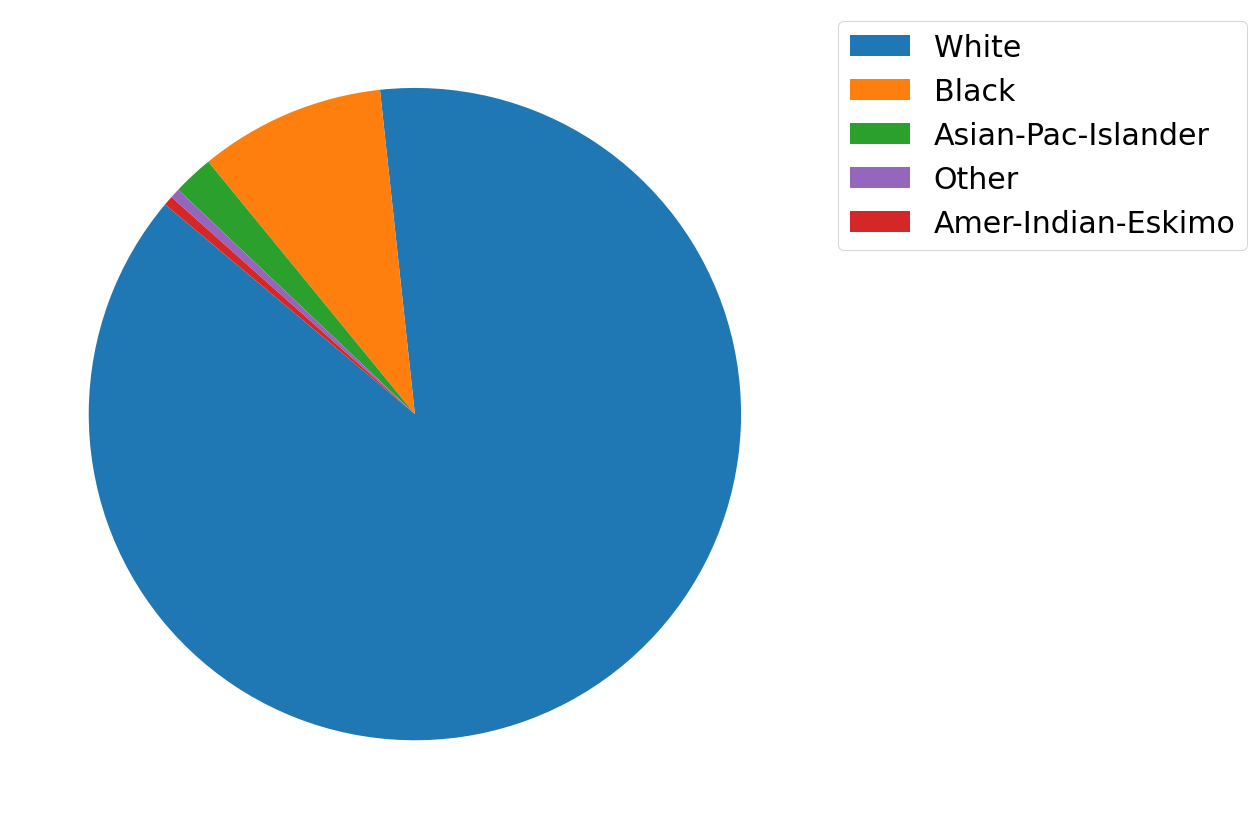} &  \includegraphics[width=0.24\textwidth]{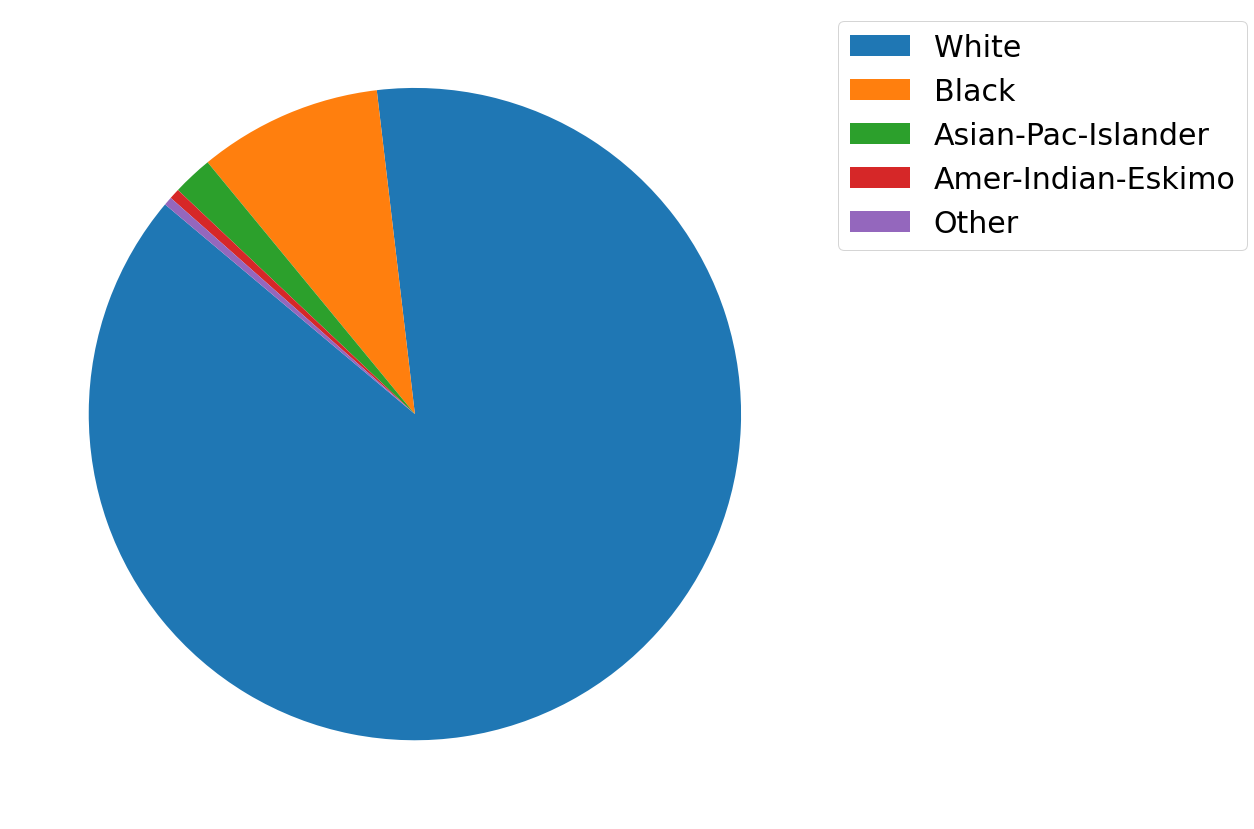} &  \includegraphics[width=0.24\textwidth]{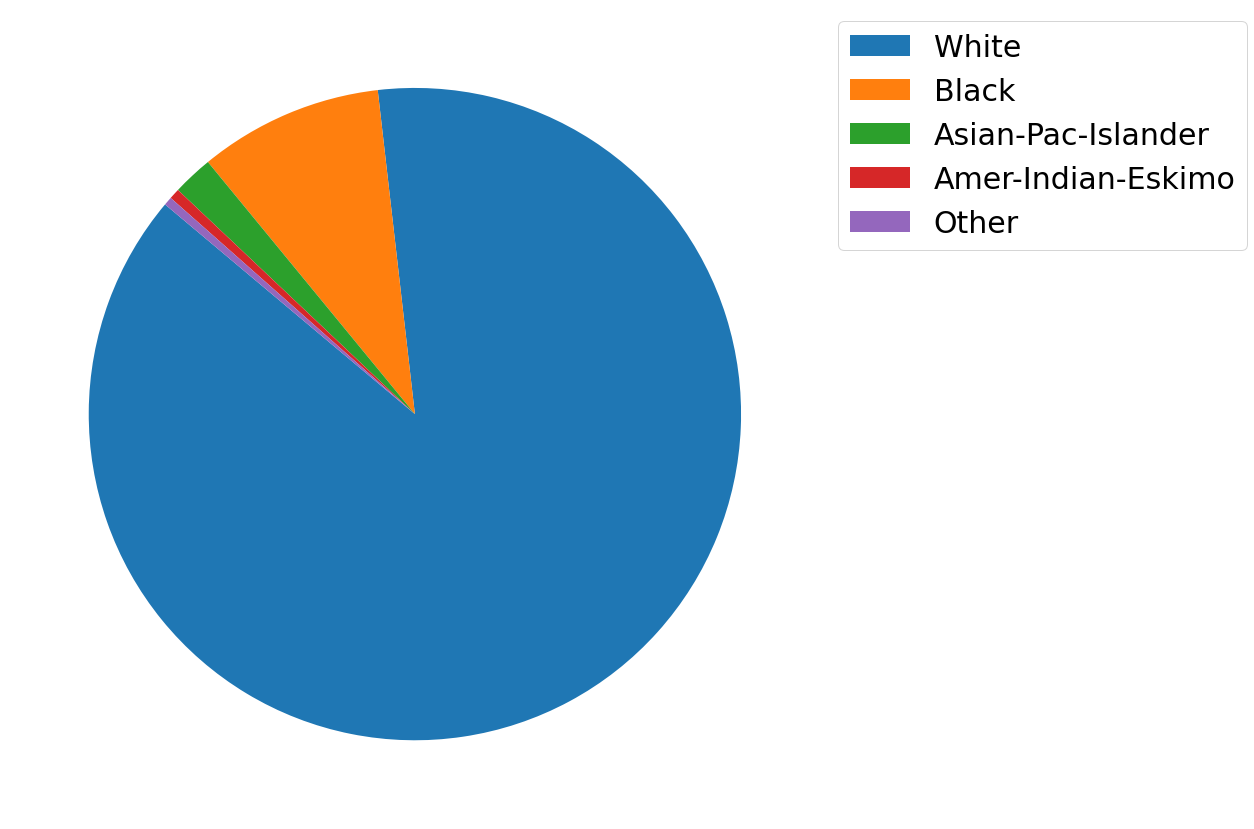}  \\
  (5) Original race &   (6) 20,000 race case &   (7) 100,000 race case &   (8) 150,000 race case \\
  
   \includegraphics[width=0.22\textwidth]{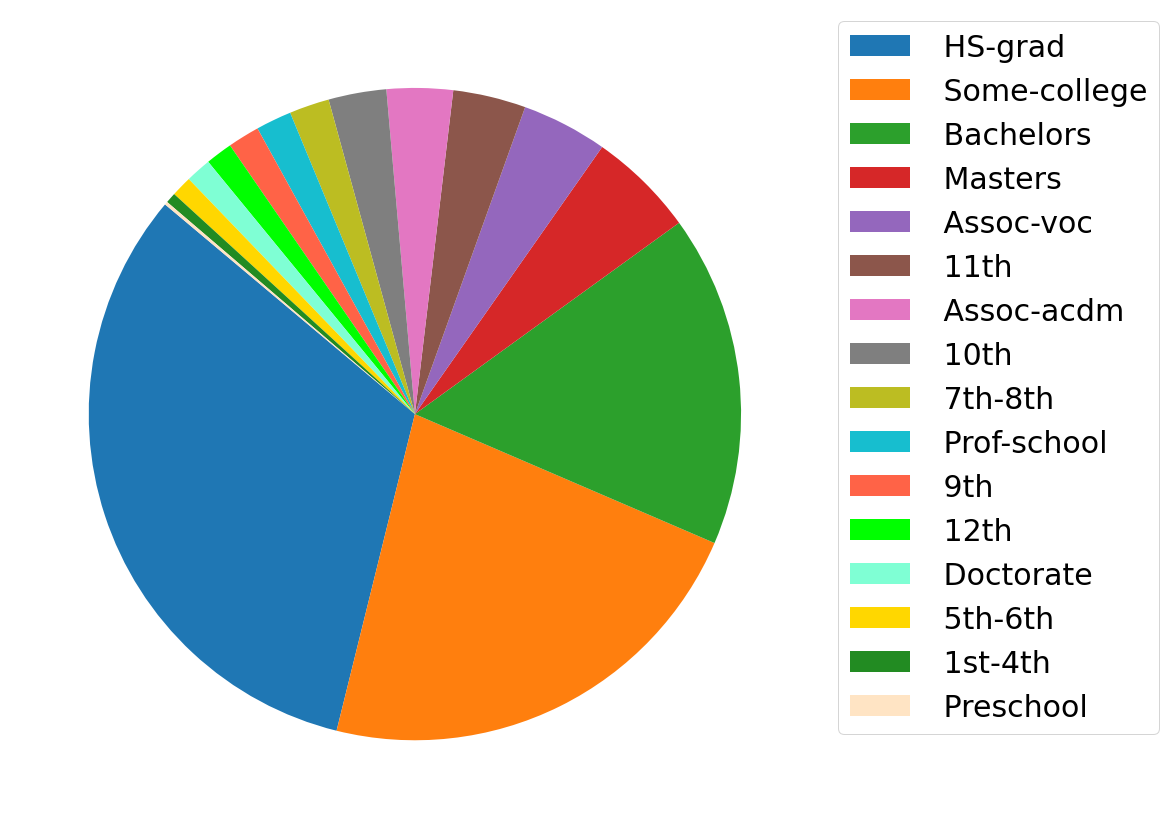} &   \includegraphics[width=0.22\textwidth]{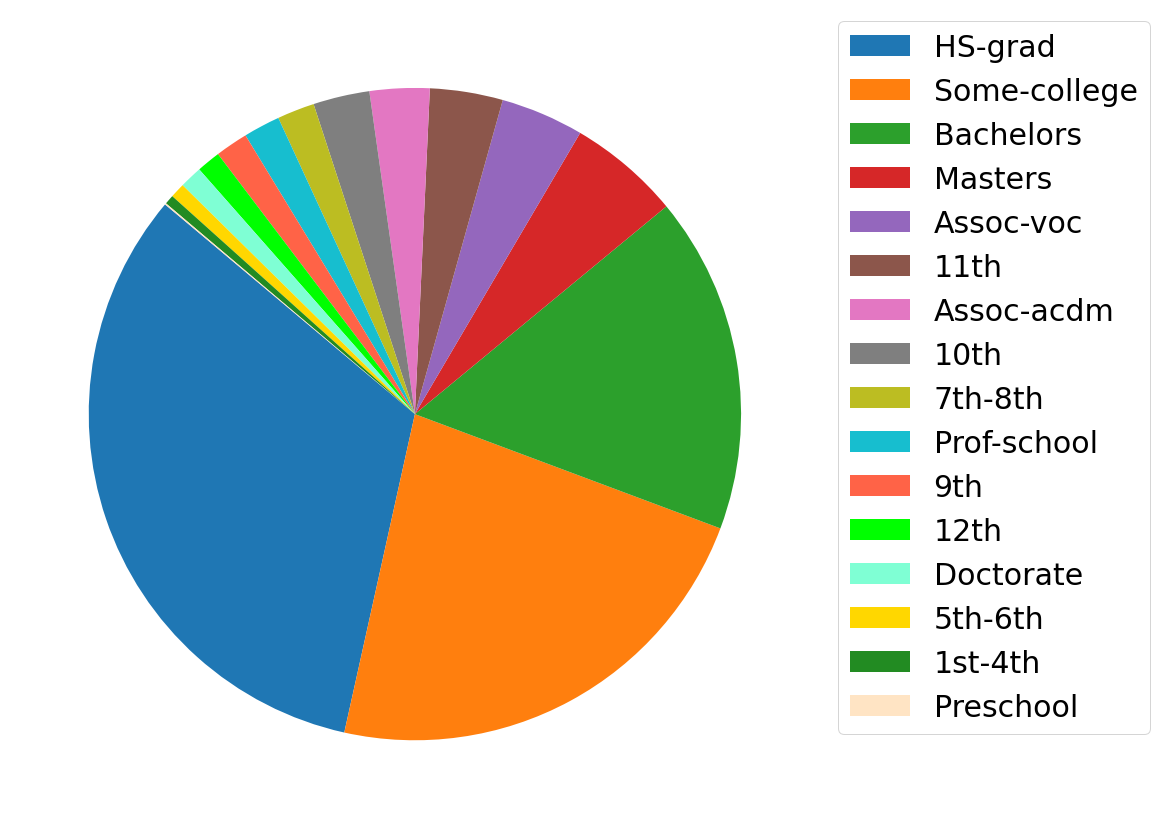} &  \includegraphics[width=0.22\textwidth]{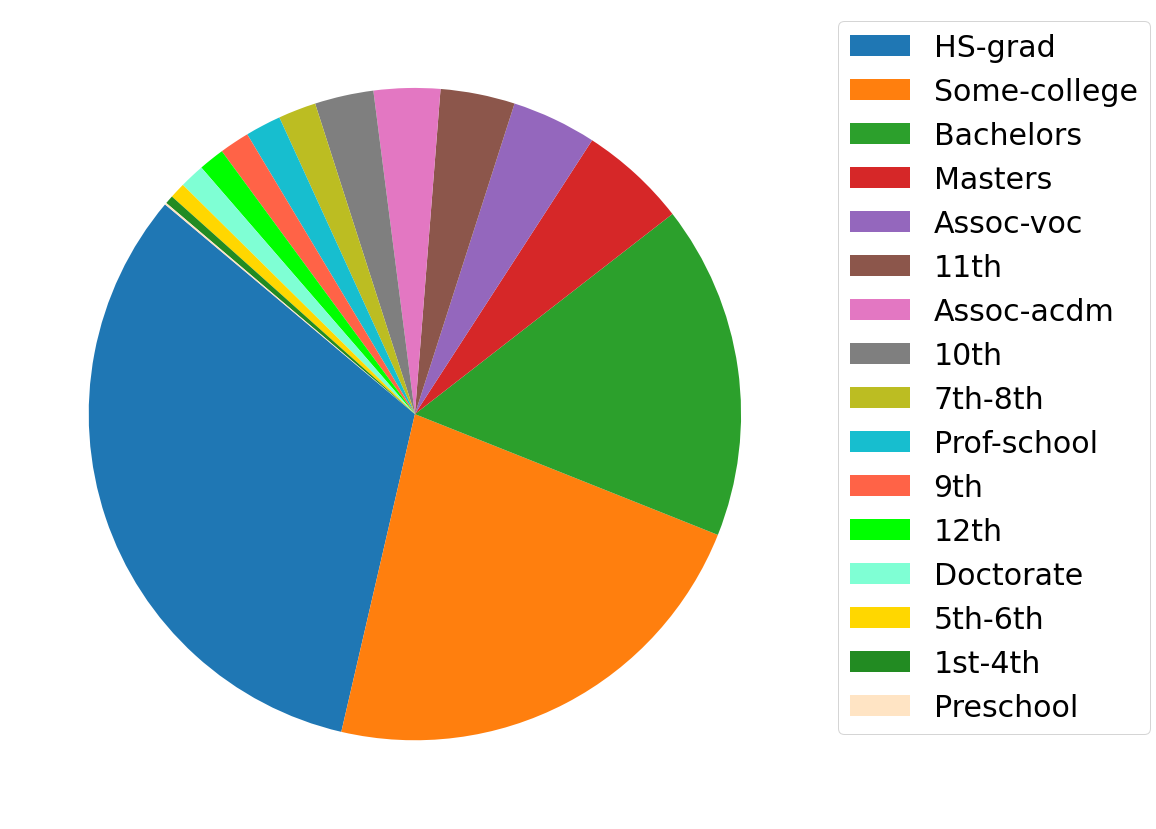} &  \includegraphics[width=0.22\textwidth]{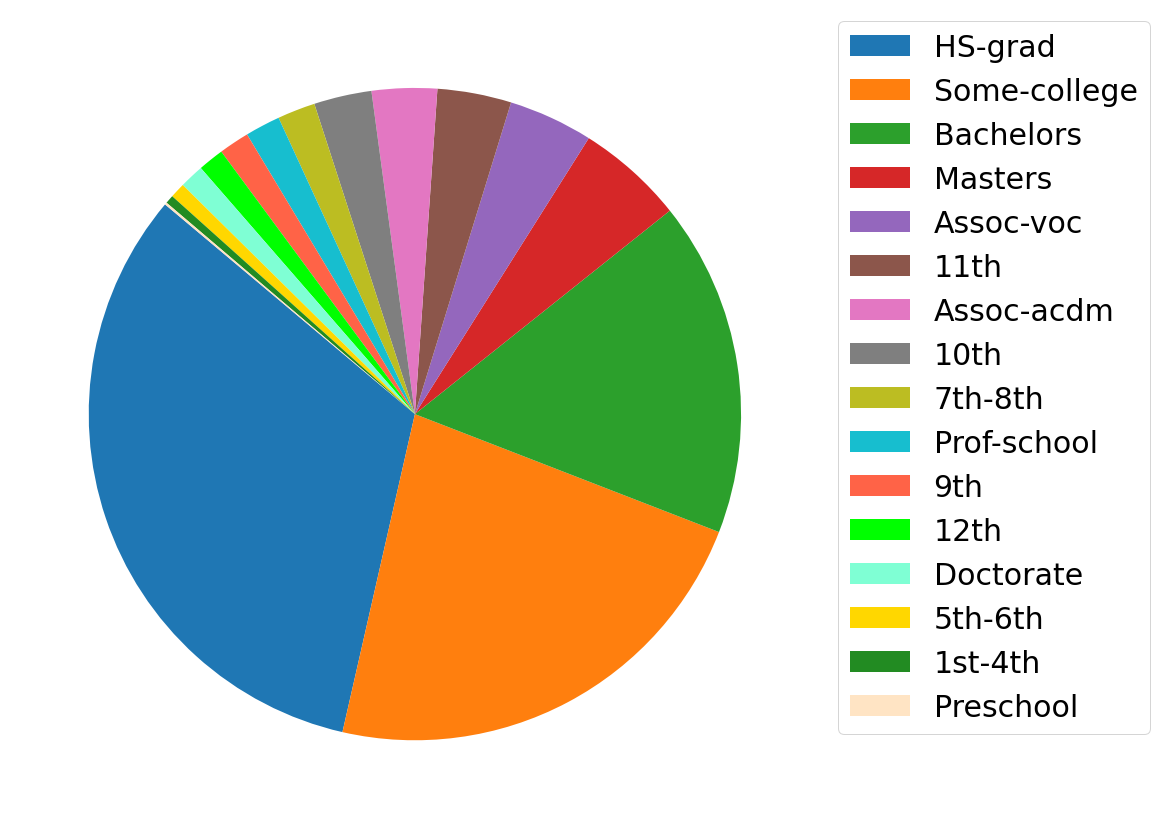}  \\
  (9) Original education &   (10) 20,000 education case &   (11) 100,000 education case  &   (12) 150,000 education case  \\
  
   \includegraphics[width=0.22\textwidth]{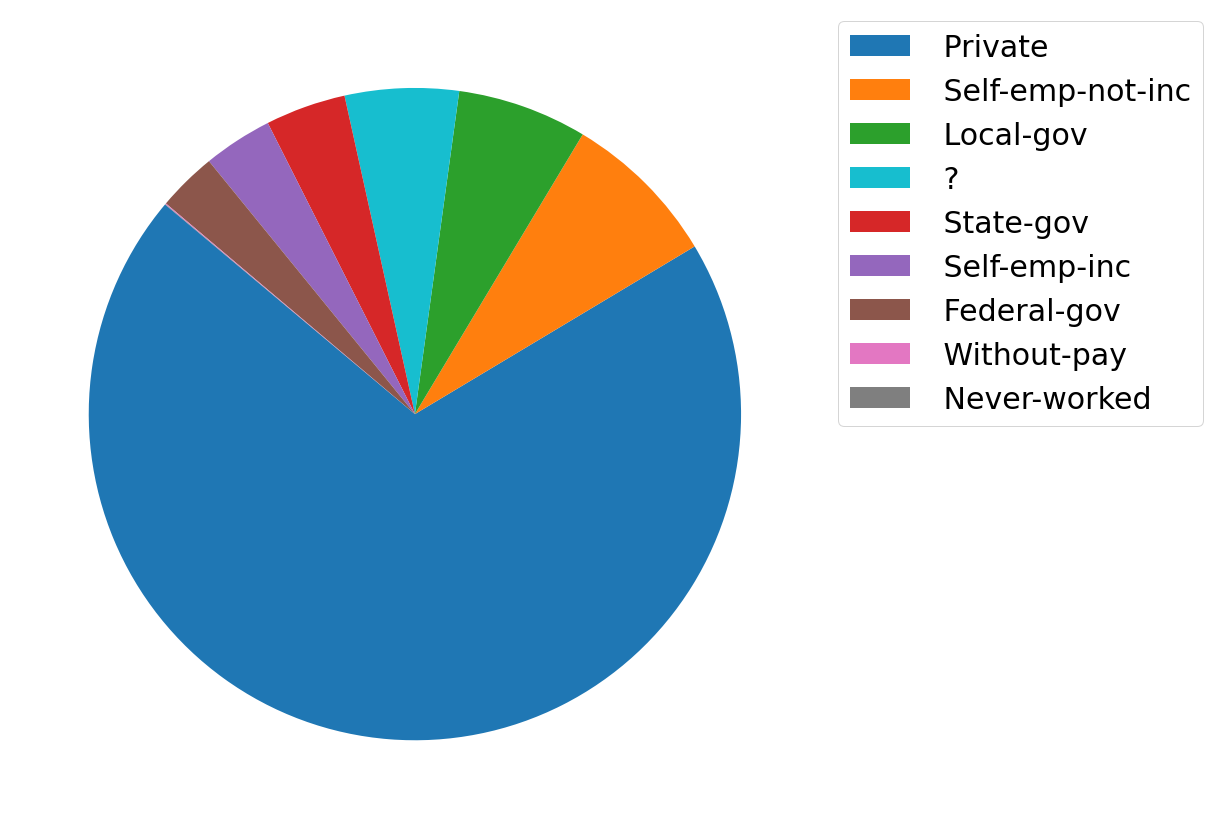} &   \includegraphics[width=0.22\textwidth]{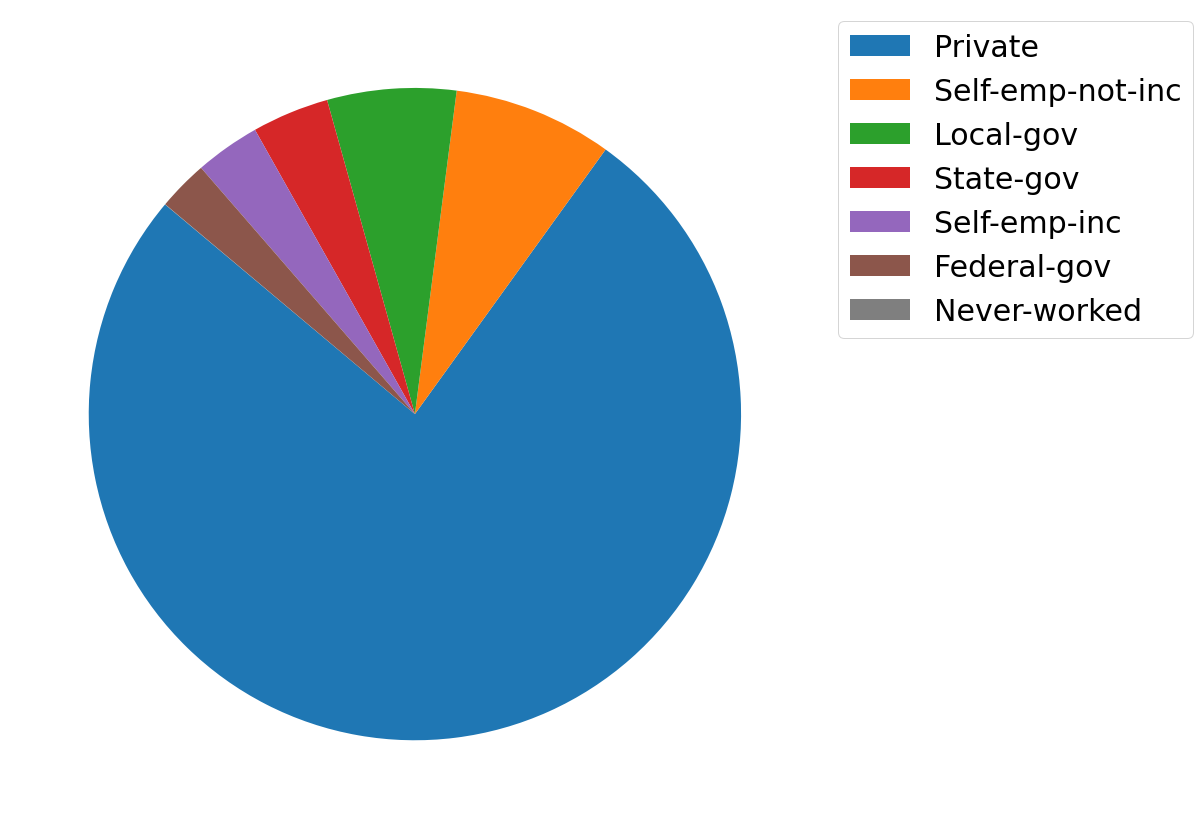} &  \includegraphics[width=0.22\textwidth]{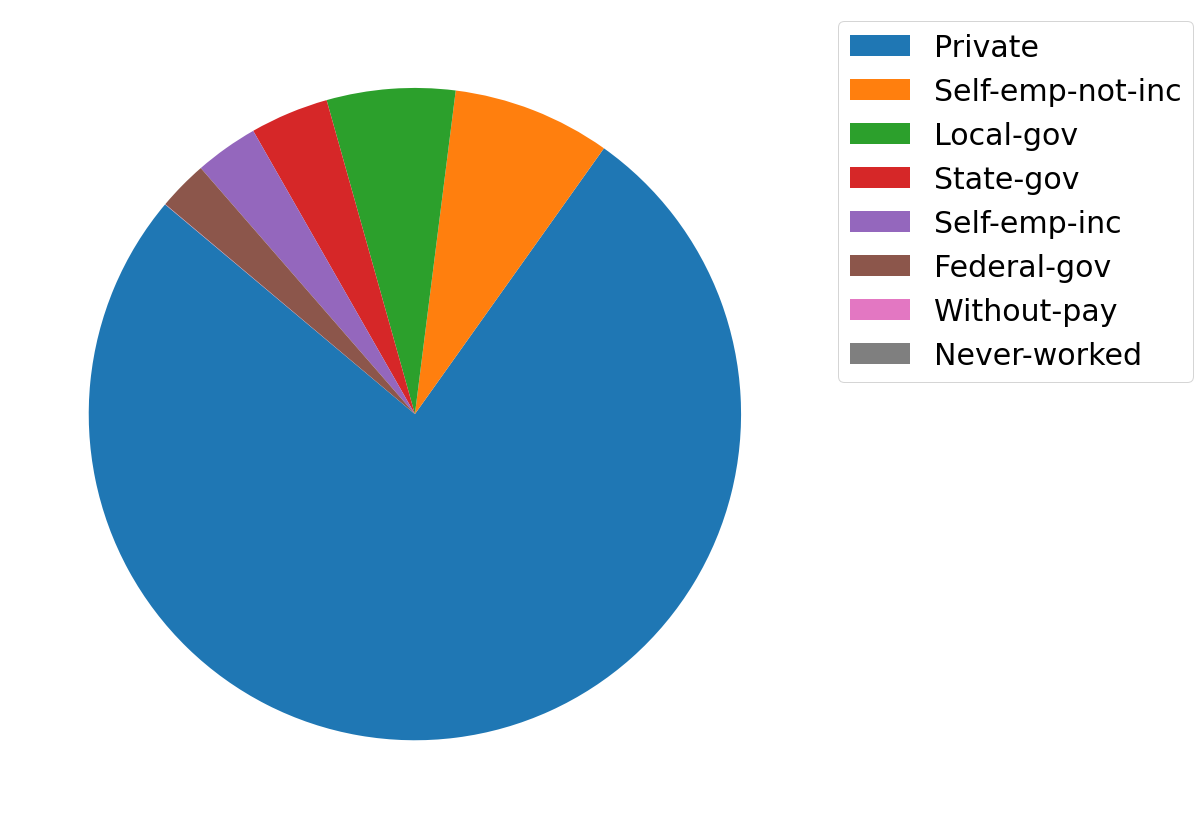} &  \includegraphics[width=0.22\textwidth]{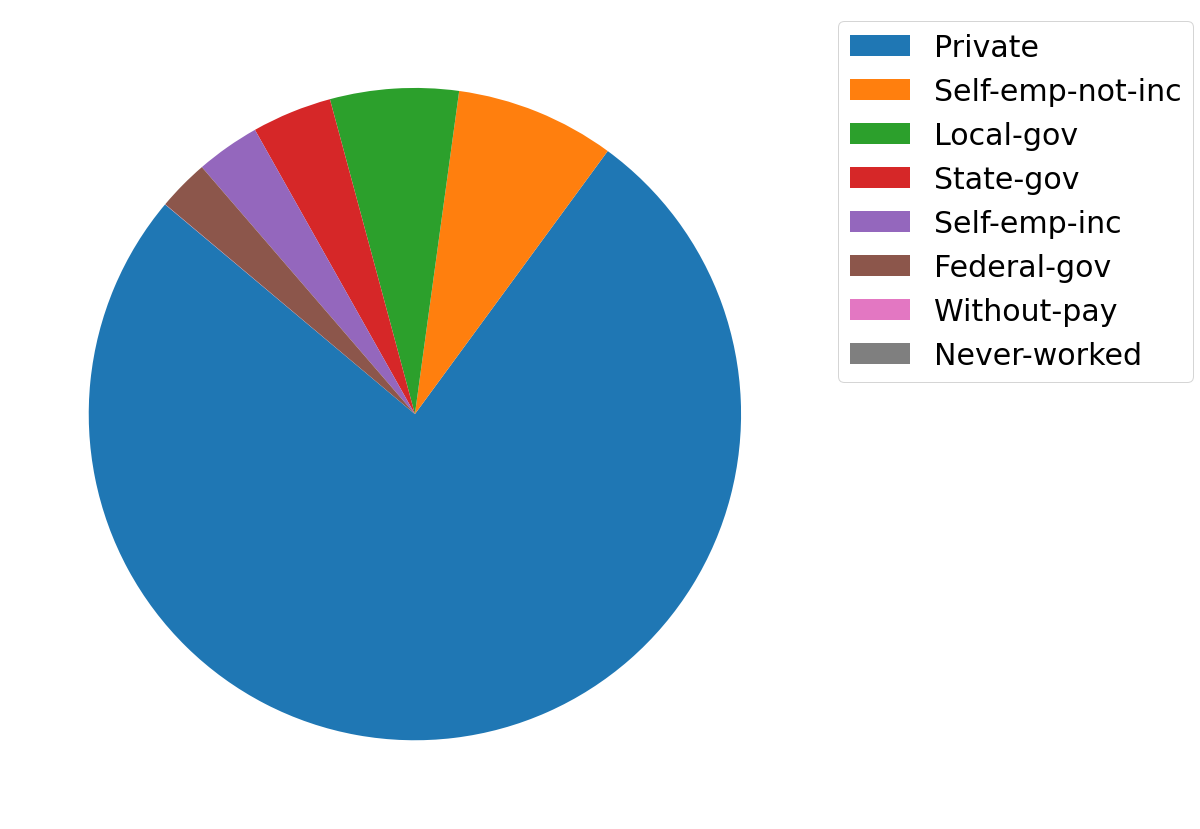}  \\
  (13) Original work class &   (14) 20,000 work class case &   (15) 100,000 work class case &   (16) 150,000 work class case \\
  
     \includegraphics[width=0.22\textwidth]{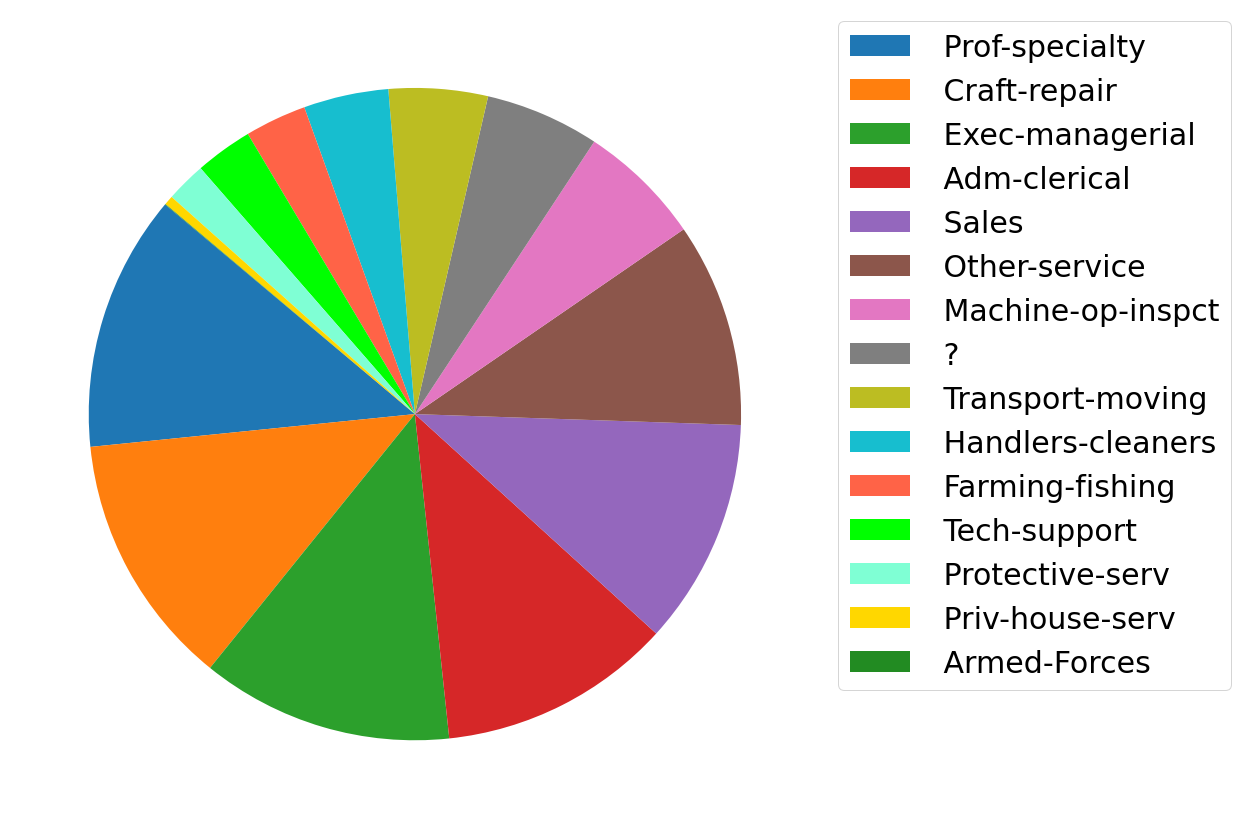} &   \includegraphics[width=0.22\textwidth]{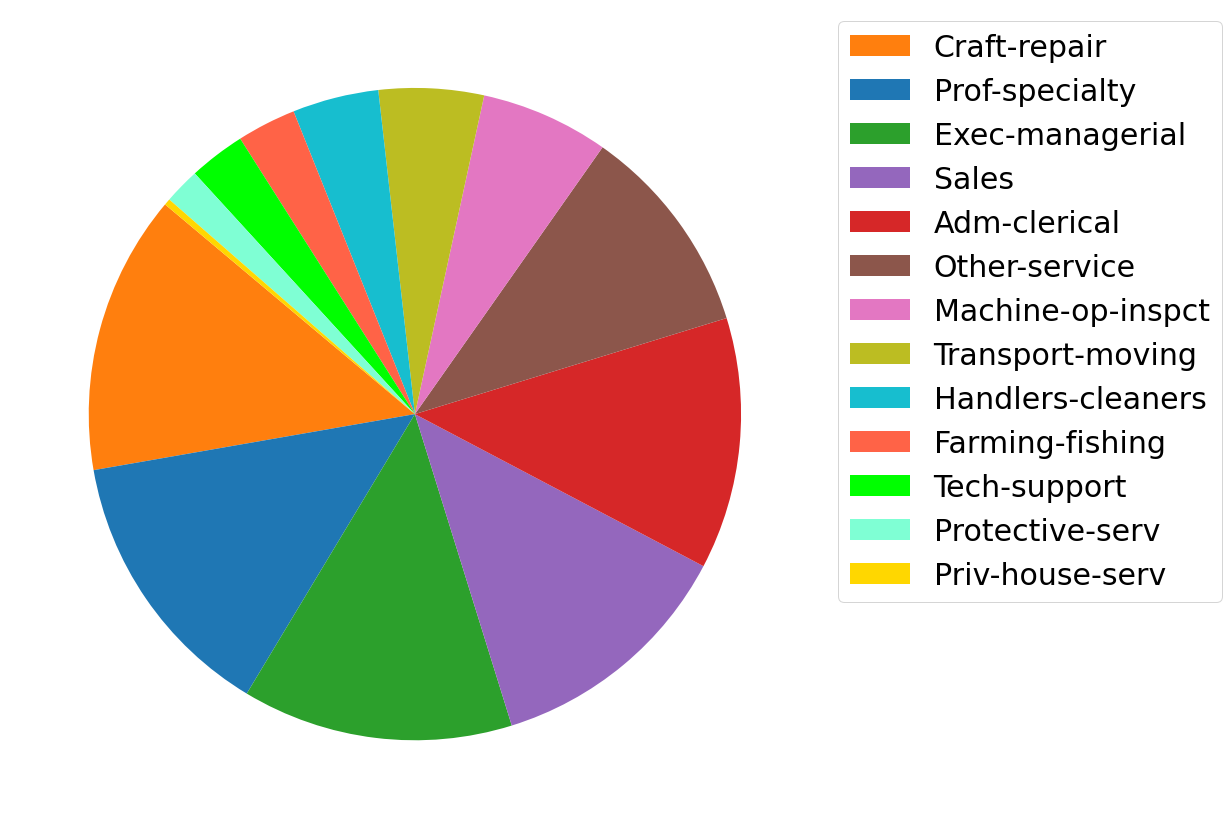} &  \includegraphics[width=0.22\textwidth]{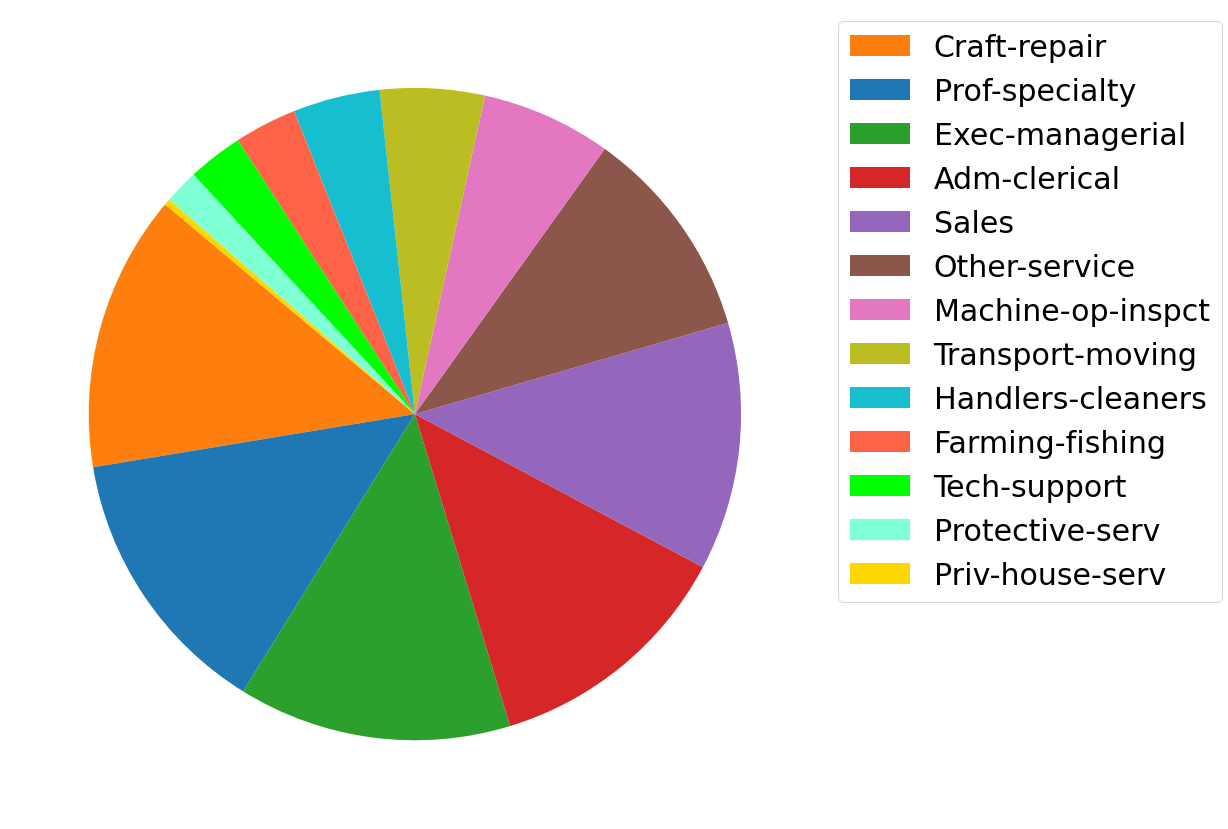} &  \includegraphics[width=0.22\textwidth]{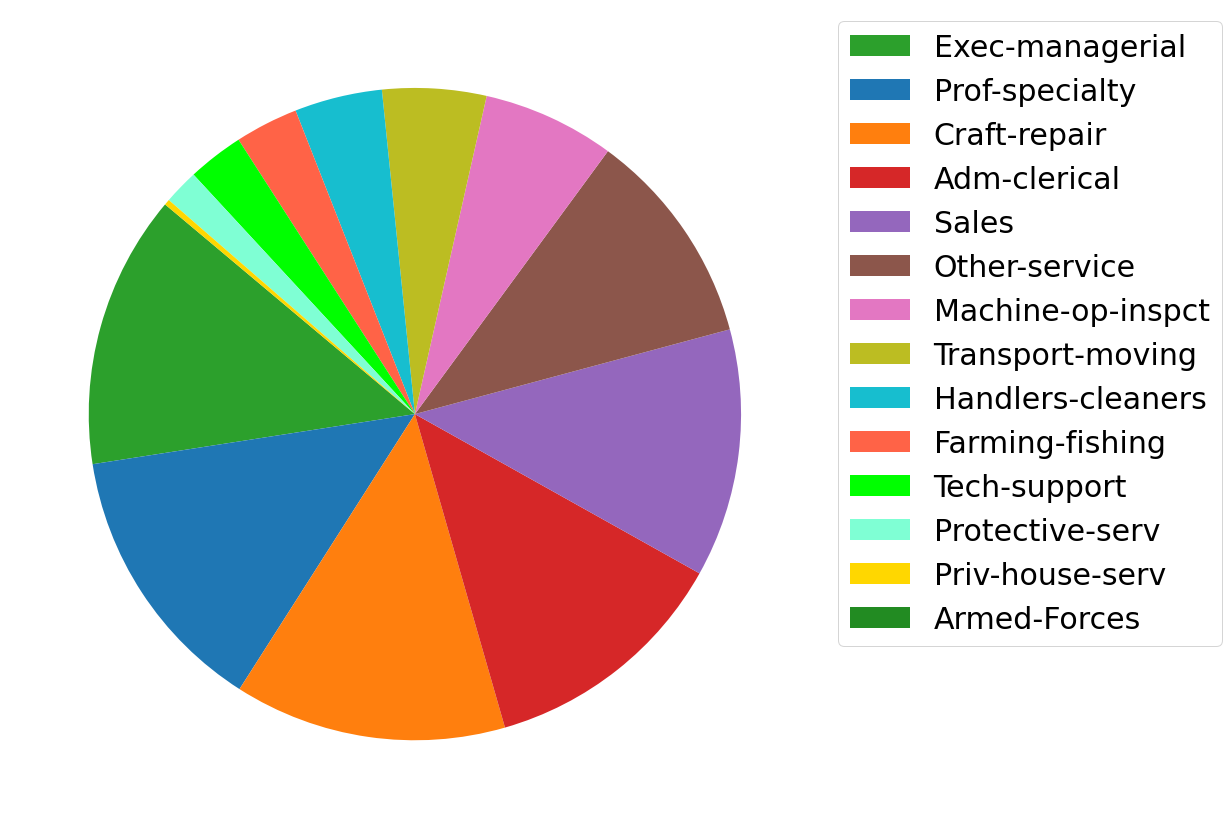}  \\
  (17) Original occupation &   (18) 20,000 occupation case &   (19) 100,000 occupation case &   (20) 150,000 occupation case \\
\end{tabular}
\caption{Attribute distribution comparison between original data and synthetic data. It includes five attributes, namely, sex, race, education, work class, and occupation. Specifically, the attribute value ``?" refers to missing data for the attributes. In addition, there are three cases for synthetic data including 20,000 synthetic samples, 100,000 synthetic samples,, and 150,000 synthetic samples for these attributes.}
\label{fig_synthetic}
\end{figure*}

\textbf{Synthetic Dataset:} Tab-DDPM was employed to generate synthetic samples to implement data augmentation, enhancing both AI fairness and classification performance. Synthetic data was added to the training set in varying sample sizes: $20,000$, $100,000$, and $150,000$ samples. Figure~\ref{fig_synthetic} compares the attribute distributions between the original Adult Income dataset and the synthetic datasets. The distributions of synthetic data closely resemble the original data across all sample sizes. Furthermore, the synthetic data is free of missing values, improving overall data quality. These observations suggest that synthetic data is promising for data augmentation, particularly in terms of maintaining data quality.

\begin{table}[ht]
	\caption{Fairness ranges of various evaluation metrics. } 
        \begin{center}
                \begin{tabular}{|cc|}
                    \hline \textbf{Metrics}  & \textbf{Fairness Ranges}  \\ 
                    \hline
                    	  SPD		&  $-0.10 \le fair \le 0.10$		 \\ 
                            EOD			&  $0.10 \le fair \le 0.10$			\\ 
                            AOD			&  $-0.10 \le fair \le 0.10$		  \\ 
                            DI			&  $0.80 \le fair \le 1.20$			\\ 
                            TI			&  $0 \le fair \le 0.25$		 \\                        
                     \hline
                \end{tabular}
        \end{center}
         \label{fairnes_metrics}
\end{table}

\begin{table*}[ht]
	\caption{Performance comparison between all outputs before reweighting through one classification metric BA and fairness metrics including SPD, AOD, DI, EOD and TI on Adult income dataset regarding the protected attribute $Race$. It presents the performances of adding different numbers of synthetic samples to the original data, including $20,000$, $100,000$, and $150,000$. } 
        \begin{center}
                \begin{tabular}{|l|cccccc|l|cccccc|}
                \hline \multicolumn{7}{|c|}{Performance before reweighting samples (original)} & \multicolumn{7}{|c|}{Performance after reweighting samples (original)}\\ \hline
                    \hline \textbf{Model}  & BA & SPD & AOD & DI & EOD & TI & \textbf{Model}  & BA & SPD & AOD & DI & EOD & TI \\ \hline
                    	  DT					&  0.7426		&  $-$0.2416		&  $-$0.1959		&  0.4196 		&  $-$0.2026		&  0.1130	& DT					&   1.0		&  $-$0.1066		&  0.0		&  0.5863 		&  0.0		&  0.0	 \\
                            GNB			&  0.7416		&  $-$0.2952		&  $-$0.2623		&  0.3252 		&  $-$0.2872		&  0.1111	& GNB			& 0.7432		&  $-$0.1147		&  $-$0.0252		&  0.7379 		&  0.0310		&  0.1058\\
                            KNN			&  0.7390		&  $-$0.1904		&  $-$0.1409		&  0.4882 		&  $-$0.1416		&  0.1207 & KNN			&  0.7390		&  $-$0.1904		&  $-$0.1409       	&  0.4882		&  $-$0.1416		&  0.1207  \\
                            LR			&  0.7437		&  $-$0.2435		&  $-$0.1966		&  0.4122 		&  $-$0.2020		&  0.1129 & LR			&  0.7311		&  $-$0.0523		&  0.0419		&  0.8508 		&  0.1083		&  0.1247	\\
                            RF				&  0.7471		&  $-$0.2014		&  $-$0.1336		&  0.5380		&  $-$0.1097		&  0.1066 &  RF				& 0.7447		&  $-$0.1072		&  $-$0.0201		&  0.7449		&  0.0321		&  0.1081  \\ 
                         
                     \hline
                \hline \multicolumn{7}{|c|}{Performance before reweighting samples (20,000)} & \multicolumn{7}{|c|}{Performance after reweighting samples (20,000)}\\ \hline
                    \hline \textbf{Model}  & BA & SPD & AOD & DI & EOD & TI & \textbf{Model}  & BA & SPD & AOD & DI & EOD & TI\\ \hline
                    	  DT					&  0.7441		&  $-$0.2182		&  $-$0.1763		&  0.4791 		&  $-$0.1791		&  0.1117	 &  DT					&  1.0		&  $-$0.0913		&  0.0		&  0.6444 		&  0.0		&  0.0\\
                            GNB			&  0.7373		&  $-$0.3107		&  $-$0.2975		&  0.2965 		&  $-$0.3417		&  0.1129	& GNB			&  0.7354		&  $-$0.0984		&  $-$0.0358		&  0.7574 		&  $-$0.0051		&  0.1151\\
                            KNN			&  0.7303		&  $-$0.1021		&  $-$0.0262		&  0.7544 		&  0.0262		&  0.1156 & KNN			&  0.7293		&  $-$0.1067		&  $-$0.0302		&  0.7461		&  0.0236		&  0.1153\\
                            LR			&  0.7432		&  $-$0.2157		&  $-$0.1705		&  0.5099 		&  $-$0.1678		&  0.1082	& LR			&   0.7415		&  $-$0.0446		&  0.0204		&  0.8978		&  0.0526		&  0.1052\\
                            RF				&  0.7440		&  $-$0.1414		&  $-$0.0802		&  0.6776		&  $-$0.0539		&  0.1063 & RF				&  0.7440		&  $-$0.0946		&  $-$0.0243		&  0.7842		&  0.0157		&  0.1051 \\ 

                     \hline 
                      \hline \multicolumn{7}{|c|}{Performance before reweighting samples (100,000)} & \multicolumn{7}{|c|}{Performance after reweighting samples (100,000)} \\ \hline
                    \hline \textbf{Model}  & BA & SPD & AOD & DI & EOD & TI & \textbf{Model}  & BA & SPD & AOD & DI & EOD & TI \\ \hline
                    	  DT					&  0.7496		&  $-$0.1260		&  $-$0.0573		&  0.7148 		&  $-$0.0313		&  0.1020	& DT					&   1.0		&  $-$0.1031		&  0.0		&  0.5958 		&  0.0		&  0.0 \\
                            GNB			&  0.7345		&  $-$0.3976		&  $-$0.3794		&  0.2321 		&  $-$0.4205		&  0.1003	& GNB			&   0.7440		&  $-$0.1133		&  $-$0.0455		&  0.7462 		&  $-$0.0191		&  0.1032\\
                            KNN			&  0.7233		&  $-$0.1619		&  $-$0.1009		&  0.5979		&  $-$0.0778		&  0.1214 & KNN			&  0.7234		&  $-$0.1602		&  $-$0.0980		&  0.6022		&  $-$0.0732		&  0.1213 \\
                            LR			&  0.7504		&  $-$0.2324		&  $-$0.1788		&  0.4756 		&  $-$0.1741		&  0.1039	& LR			&  0.7487		&  $-$0.1044		&  $-$0.0388		&  0.7554		&  $-$0.0170		&  0.1045\\
                            RF				&  0.7503		&  $-$0.1699		&  $-$0.1084		&  0.6144		&  $-$0.0926		&  0.1029 & RF				&  0.7489	&  $-$0.1092		&  $-$0.0427		&  0.7525		&  $-$0.0196		&  0.1020 \\ 
                         
                     \hline
                      \hline \multicolumn{7}{|c|}{Performance before reweighting samples (150,000)} & \multicolumn{7}{|c|}{Performance after reweighting samples (150,000)}\\ \hline
                    \hline \textbf{Model}  & BA & SPD & AOD & DI & EOD & TI & \textbf{Model}  & BA & SPD & AOD & DI & EOD & TI \\ \hline
                    	  DT					&  0.7473		&  $-$0.2262		&  $-$0.1895		&  0.4790 		&  $-$0.2081		&  0.1066	& DT					&  1.0		&  $-$0.1020		&  0.0		&  0.5995 		&  0.0		&  0.0 \\
                            GNB			&  0.7262		&  $-$0.4150		&  $-$0.3996		&  0.2226 		&  $-$0.4418		&  0.1006 & GNB					&   0.7296		&  $-$0.0906		&  $-$0.0344		&  0.8225 		&  $-$0.0202		&  0.0955	\\
                            KNN			&  0.7382		&  $-$0.1237		&  $-$0.0797		&  0.6983 		&  $-$0.0858		&  0.1126 & KNN					&  0.7382		&  $-$0.1237		&  $-$0.0797		&  0.6983		&  $-$0.0858		&  0.1126\\
                            LR			&  0.7468		&  $-$0.2180		&  $-$0.1782		&  0.5007 		&  $-$0.1922		&  0.1062	& LR					&  0.7414		&  $-$0.0127		&  0.0340		&  0.9709		&  0.0299		&  0.1037\\
                            RF				&  0.7476		&  $-$0.1862		&  $-$0.1379		&  0.5713		&  $-$0.1399		&  0.1056 & RF					&  0.7461	&  $-$0.0895		&  $-$0.0300		&  0.7939		&  $-$0.0158		&  0.1040 \\ \hline
                         
                \end{tabular}
        \end{center}
         \label{tab_adult_race}
\end{table*}

\begin{table*}[ht]
	\caption{Performance comparison between all outputs before reweighting through one classification metric BA and fairness metrics including SPD, AOD, DI, EOD and TI on Adult income dataset regarding the protected attribute $Sex$. } 
        \begin{center}
                \begin{tabular}{|l|cccccc|l|cccccc|}
                \hline \multicolumn{7}{|c|}{Performance before reweighting samples (original)} & \multicolumn{7}{|c|}{Performance after reweighting samples (original)}\\ \hline
                    \hline \textbf{Model}  & BA & SPD & AOD & DI & EOD & TI & \textbf{Model}  & BA & SPD & AOD & DI & EOD & TI \\ \hline
                    	  DT					&  0.7426		&  $-$0.3608		&  $-$0.3204		&  0.2785 		&  $-$0.3775		&  0.1130	& DT				  &  1.0		&  $-$0.1910		&  0.0		&  0.3740 		&  0.0		&  0.0	 \\
                            GNB			&  0.7416		&  $-$0.3353		&  $-$0.2805		&  0.3369 		&  $-$0.3184		&  0.1111	& GNB			&  0.7209		&  $-$0.0861		&  0.0073		&  0.7997 		&  0.0203		&  0.1192\\
                            KNN			&  0.7390		&  $-$0.3983		&  $-$0.4075		&  0.1616 		&  $-$0.5311		&  0.1207 & KNN			&  0.7390		&  $-$0.3983		&  $-$0.4075       &  0.1616		&  $-$0.5311		&  0.1207   \\
                            LR			&  0.7437		&  $-$0.3580		&  $-$0.3181		&  0.2794 		&  $-$0.3769		&  0.1129 & LR			&  0.7134		&  $-$0.0705		&  0.0188		&  0.7785 		&  0.0293		&  0.1401	\\
                            RF				&  0.7471		&  $-$0.3777		&  $-$0.3292		&  0.2884		&  $-$0.3763		&  0.1066 &  RF				&  0.7271		&  $-$0.1386		&  $-$0.0638		&  0.7220		&  $-$0.0774		&  0.1065  \\ 
                         
                     \hline
                \hline \multicolumn{7}{|c|}{Performance before reweighting samples (20,000)} & \multicolumn{7}{|c|}{Performance after reweighting samples (20,000)}\\ \hline
                    \hline \textbf{Model}  & BA & SPD & AOD & DI & EOD & TI & \textbf{Model}  & BA & SPD & AOD & DI & EOD & TI\\ \hline
                    	  DT					&  0.7441		&  $-$0.2182		&  $-$0.1763		&  0.4791 		&  $-$0.1791		&  0.1117	 &  DT					&  1.0		&  $-$0.1957		&  0.0		&  0.3657 		&  0.0		&  0.0\\
                            GNB			&  0.7373		&  $-$0.3600		&  $-$0.3072		&  0.3044 		&  $-$0.3461		&  0.1129	& GNB			&  0.7143		&  $-$0.0984		&  $-$0.0215		&  0.7791 		&  $-$0.0303		&  0.1205\\
                            KNN			&  0.7303		&  $-$0.2969		&  $-$0.2315		&  0.4052 		&  $-$0.2528		&  0.1156 & KNN			&  0.7293		&  $-$0.3028		&  $-$0.2366		&  0.4005		&  $-$0.2556		&  0.1153 \\
                            LR			&  0.7432		&  $-$0.3905		&  $-$0.3344		&  0.2753 		&  $-$0.3699		&  0.1082	& LR			&  0.7173		&  $-$0.0100		&  $-$0.0244		&  0.8020		&  $-$0.0369		&  0.1071 \\
                            RF				&  0.7440		&  $-$0.3993		&  $-$0.3379		&  0.2740		&  $-$0.3656		&  0.1063 & RF				&  0.7188	&  $-$0.1065		&  $-$0.0318		&  0.7920		&  $-$0.0458		&  0.1053  \\ 

                     \hline 
                      \hline \multicolumn{7}{|c|}{Performance before reweighting samples (100,000)} & \multicolumn{7}{|c|}{Performance after reweighting samples (100,000)} \\ \hline
                    \hline \textbf{Model}  & BA & SPD & AOD & DI & EOD & TI & \textbf{Model}  & BA & SPD & AOD & DI & EOD & TI \\ \hline
                    	  DT					&  0.7496		&  $-$0.4149		&  $-$0.3538		&  0.2580 		&  $-$0.3871		&  0.1020	& DT					&  1.0		&  $-$0.2029		&  0.0		&  0.3394 		&  0.0		&  0.0 \\
                            GNB			&  0.7345		&  $-$0.4182		&  $-$0.3477		&  0.3047 		&  $-$0.3588		&  0.1003	& GNB			&  0.7195		&  $-$0.1084		&  $-$0.0220		&  0.7711 		&  $-$0.0242		&  0.1123\\
                            KNN			&  0.7233		&  $-$0.3791		&  $-$0.3225		&  0.2478 		&  $-$0.3491		&  0.1214 & KNN			&  0.7234		&  $-$0.3795		&  $-$0.3229		&  0.2477		&  $-$0.3495		&  0.1213 \\
                            LR			&  0.7504		&  $-$0.4016		&  $-$0.3460		&  0.2601 		&  $-$0.3883		&  0.1039	& LR			&  0.7249		&  $-$0.1089		&  $-$0.0196		&  0.7639		&  $-$0.0216		&  0.1124\\
                            RF				&  0.7503		&  $-$0.4047		&  $-$0.3447		&  0.2631		&  $-$0.3806		&  0.1029 & RF				&  0.7192	&  $-$0.0633		&  0.0257		&  0.8602		&  $-$0.0240		&  0.1137 \\ 
                         
                     \hline
                      \hline \multicolumn{7}{|c|}{Performance before reweighting samples (150,000)} & \multicolumn{7}{|c|}{Performance after reweighting samples (150,000)}\\ \hline
                    \hline \textbf{Model}  & BA & SPD & AOD & DI & EOD & TI & \textbf{Model}  & BA & SPD & AOD & DI & EOD & TI \\ \hline
                    	  DT					&  0.7473		&  $-$0.4031		&  $-$0.3579		&  0.2494 		&  $-$0.4098		&  0.1066	& DT					&  1.0		&  $-$0.1956		&  0.0		&  0.3600 		&  0.0		&  0.0\\
                            GNB			&  0.7262		&  $-$0.3525		&  $-$0.2867		&  0.4090 		&  $-$0.3007		&  0.1006 & GNB					&  0.7189		&  $-$0.0964		&  $-$0.0245		&  0.7967 		&  $-$0.0433   &  0.1118	\\
                            KNN			&  0.7382		&  $-$0.3494		&  $-$0.2938		&  0.3124 		&  $-$0.3296		&  0.1126 & KNN					&  0.7382		&  $-$0.3493		&  $-$0.2938		&  0.3124		&  $-$0.3296		&  0.1126 \\
                            LR			&  0.7468		&  $-$0.3961		&  $-$0.3487		&  0.2641 		&  $-$0.3977		&  0.1062	& LR					&  0.7200		&  $-$0.0976		&  $-$0.0254		&  0.8069		&  $-$0.0441		&  0.1048 \\
                            RF				&  0.7476		&  $-$0.3916		&  $-$0.3383		&  0.2725		&  $-$0.3799		&  0.1056  & RF					&  0.7230	&  $-$0.1065		&  $-$0.0289		&  0.7914		&  $-$0.0411		&  0.1032  \\ \hline
                         
                \end{tabular}
        \end{center}
         \label{tab_adult_sex}
\end{table*}

\subsection{Evaluation Metrics}

This paper utilized five evaluation metrics to determine the effectiveness of reweighting samples for mitigating bias.

Disparate Impact (DI) refers to the unintentional bias that can occur when predictions result in varying error rates or outcomes across different demographic groups, where certain attributes like race, sex, religion, and age are considered protected. This bias may arise from training models on biased data or from the model itself being discriminatory. In this study, Disparate Impact is defined as the differential effects on prediction outcomes.

\begin{equation}
DI =  \frac{p_{pup}}{p_{pp}} 
\end{equation}

where $p_{pup}$ presents the prediction probability for unprivileged samples with positive predictions, while $p_{pp}$ denotes the prediction probability for privileged samples with positive predictions. A disparate impact value approaching 0 indicates bias in favor of the privileged group, while a value greater than 1 indicates bias in favor of the unprivileged group. A value of 1 reflects perfect fairness in the predictions~\cite{feldman2015certifying}.

\textbf{Average odds difference (AOD)} measures the average difference in false positive rates (FPR) and true positive rates (TPR) between unprivileged and privileged groups. It is calculated as:

\begin{equation}
AOD =  \frac{(FPR_{up} - FPR_{p}) + (TPR_{up} - TPR_{p}) }{2} 
\end{equation}

where  $FPR_{up}$ and $FPR_{p}$ represent the False Positive Rates for unprivileged and privileged samples, respectively, while $TPR_{up}$ and $TPR_{p}$ represent the True Positive Rates for unprivileged and privileged samples. An AOD value of 0 indicates perfect fairness. A positive AOD suggests bias in favor of the unprivileged group, while a negative AOD indicates bias in favor of the privileged group.

\textbf{Statistical parity difference (SPD)} is to calculate the difference between the ratio of favorable outcomes in unprivileged and privileged groups. It is defined by

\begin{equation}
SPD =  p_{pup} - p_{pp}
\end{equation}

A score below 0 suggests benefits for the unprivileged group, while a score above 0 implies benefits for the privileged group. A score of 0 indicates that both groups receive equal benefits.

\textbf{Equal opportunity difference (EOD)} assesses whether all groups have an equal chance of benefiting from predictions. EOD focuses on the True Positive Rate (TPR), which reflects the model's ability to correctly identify positives in both unprivileged and privileged groups. It is defined as follows:

\begin{equation}
EOD =  TPR_{up} - TPR_{p}
\end{equation}

A value of 0 signifies perfect fairness. A positive value indicates bias in favor of the unprivileged group, while a negative value indicates bias in favor of the privileged group.

\textbf{Theil index (TI)} is also called the entropy index which measures both the group and individual fairness. It is defined by

\begin{equation}
TI =  \frac{1}{n}\sum_{i = 1}^{n}{ \frac{b_{i}}{\mu}ln\frac{b_{i}}{\mu}}
\end{equation}

where $b_{i} = \hat{y_{i}} - y_{i} + 1$ and $\mu$ is the average of $b_{i}$. A lower absolute value of TI value in this context would indicate a more equitable distribution of classification outcomes, while a higher absolute value suggests greater disparity.

Table~\ref{fairnes_metrics} presents the fairness ranges and levels of various evaluation metrics. If the values of evaluation metrics fall into these ranges, it indicates that the machine learning models perform classification without bias.

\subsection{Results and Discussion}

To comprehensively validate the proposed method, we conduct extensive experiments in that regard of two protect attributes, namely Race and Sex, to examine the effectiveness of bias mitigation. 

\textbf{Race: } Table~\ref{tab_adult_race} presents a performance comparison across all outputs before reweighting samples, using one classification metric, BA, and five fairness metrics—SPD, AOD, DI, EOD, and TI—on the Adult Income dataset, focusing on the protected attribute of $Race$. Before reweighting samples to mitigate bias, augmenting the training data does not appear to effectively reduce bias. For example, the SPD value for GNB increases with the addition of synthetic samples, indicating that synthetic data may actually exacerbate bias. Additionally, the performance of LR in terms of classification and bias mitigation remains relatively unchanged, as reflected in the BA, AOD, and TI values.

However, after reweighting the samples, bias is mitigated for most ML models, as indicated by improvements in SPD, AOD, and TI values. For instance, the bias in LR significantly decreases across all fairness metrics—SPD, AOD, DI, EOD, and TI. Similar trends are observed for other models like DT, GNB, and RF. Moreover, when more synthetic samples are added to the original training data, bias is further reduced, particularly in the case of the ($150,000$) sample size, as shown in the improved performance of LR and RF.

\textbf{Sex: } Table~\ref{tab_adult_sex} presents a performance comparison of all outputs before reweighting samples, using one classification metric (BA) and five fairness metrics (SPD, AOD, DI, EOD, and TI) on the Adult Income dataset, focusing on the protected attribute $Sex$. Before reweighting samples to address bias, similar patterns are observed: augmenting the training data does not effectively reduce bias, as shown consistently across ML models like DT and LR, particularly in fairness metrics such as SPD and AOD. Additionally, LR performance in both classification and bias mitigation remains largely unchanged.

However, after reweighting the samples, bias is reduced in most ML models, as indicated by improvements in SPD, AOD, and TI. For example, significant bias reductions are observed across all fairness metrics—SPD, AOD, DI, EOD, and TI—for models like LR, GNB, DT, and RF. Moreover, when additional synthetic samples are added to the original training data, bias is further mitigated, particularly in the case of GNB at the ($150,000$) sample size.

\begin{figure*}
\caption{Performance comparison via BA vs. AOD before and after reweighting samples on Adult Income dataset with respect to the protected attribute $Race$ for LR. in addition, it examines the performance for the case $150,000$ synthetic samples.}
\begin{tabular}{cc}
  \includegraphics[width=0.45\textwidth]{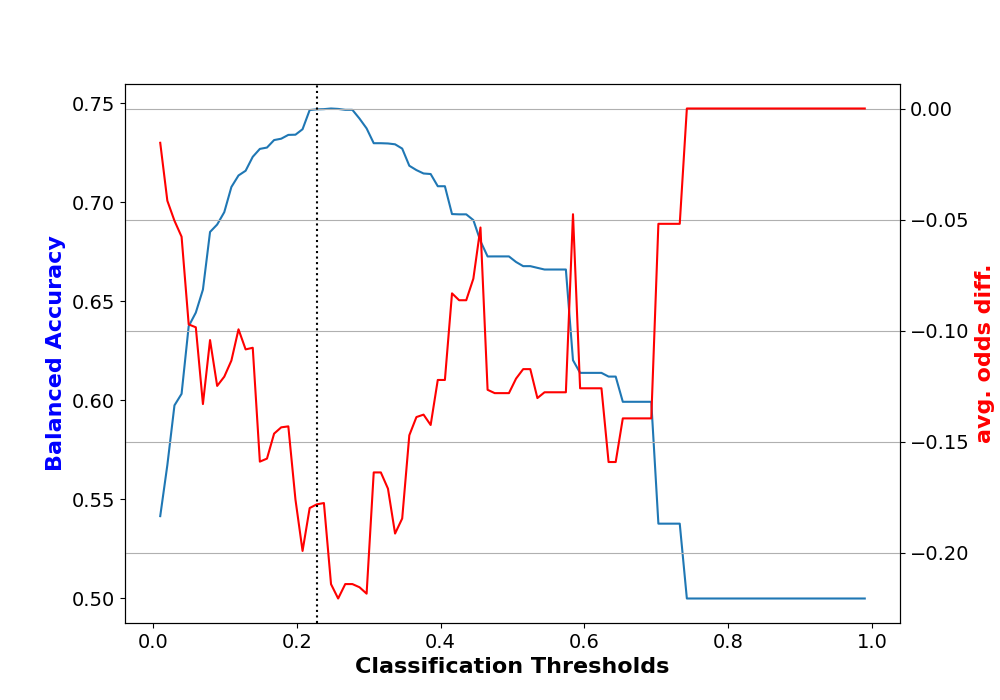} &   \includegraphics[width=0.45\textwidth]{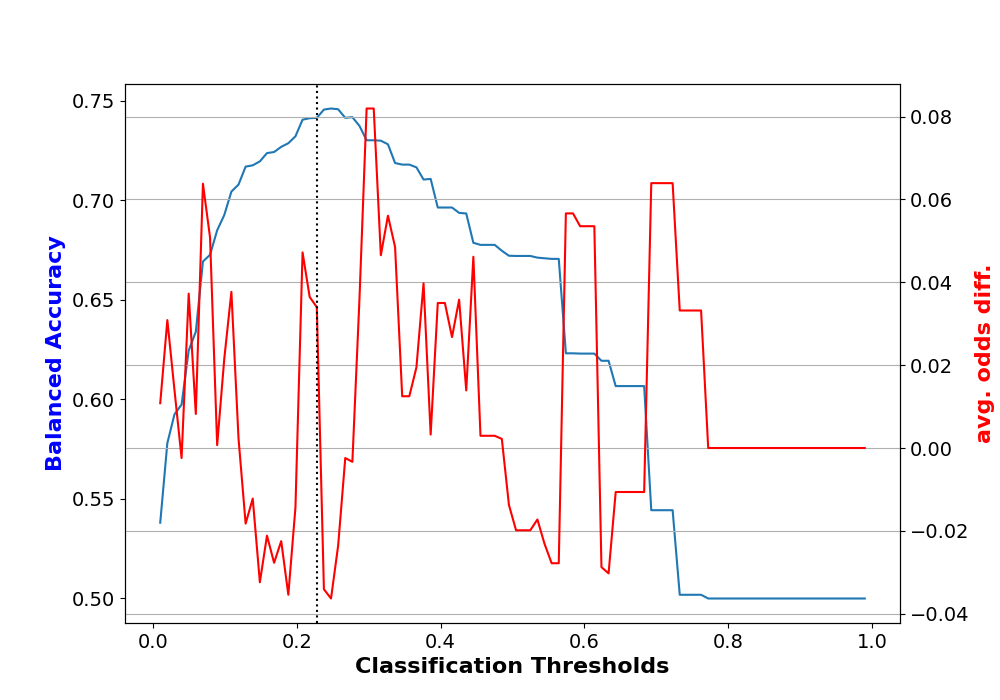} \\
(a) LR BA vs. AOD Before & (b) LR BA vs. AOD After \\[6pt]
\label{fig_LR_150000}
\end{tabular}
\end{figure*}

\begin{figure*}
\caption{Performance comparison via BA vs. AOD before and after reweighting samples on Adult Income dataset with respect to the protected attribute $Race$ for RF. in addition, it examines the performance for the case $100,000$ synthetic samples.}
\begin{tabular}{cc}
  \includegraphics[width=0.45\textwidth]{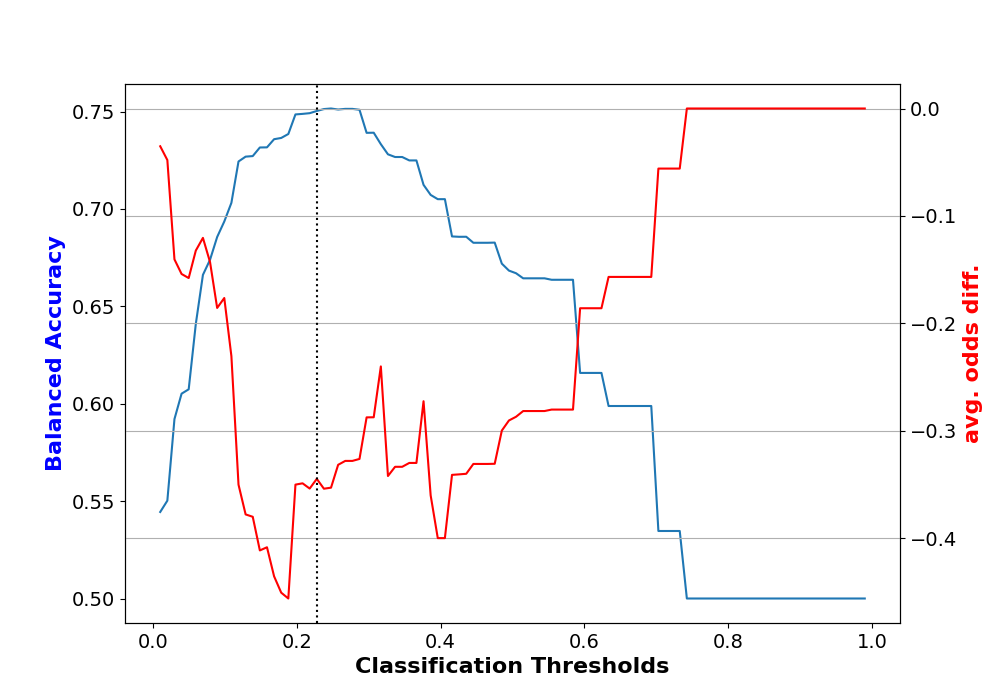} &   \includegraphics[width=0.45\textwidth]{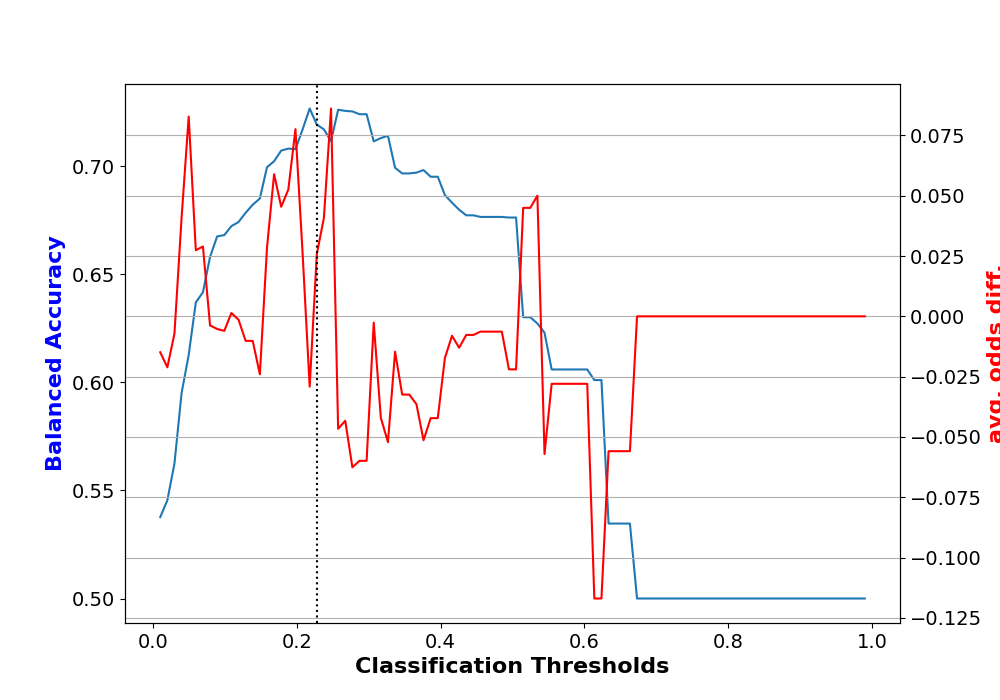} \\
(a) RF BA vs. AOD Before & (b) RF BA vs. AOD After \\[6pt]
\label{fig_RF_100000}
\end{tabular}
\end{figure*}

Additionally, we examine the correlation between classification performance (Balanced Accuracy, BA) and fairness performance (AOD), aiming for an ideal scenario where both are improved simultaneously. Figures~\ref{fig_LR_150000} and \ref{fig_RF_100000} illustrate two cases where training data is augmented with AI-generated samples—150,000 for LR and 100,000 for RF. Comparing subfigures (a) and (b) in Figure~\ref{fig_LR_150000}, it is observed that after reweighting the samples, at a classification threshold around 0.23, the AOD value for LR improves from -0.17 to 0.04, indicating enhanced fairness according to the AOD metric. However, the BA value slightly decreases from 0.75 to 0.74. Similar trends are observed in Figure~\ref{fig_RF_100000}, where at the same threshold, the AOD value for RF improves from -0.35 to 0.025, but the BA value drops from 0.75 to 0.72. These results suggest that improving fairness through augmented training with synthetic data may come at the cost of a slight reduction in classification performance. 

\section{Related Work }
\label{sec5}

\subsection{Generative Models}
Generative models have a rich history in artificial intelligence, starting in the 1950s with the development of Hidden Markov Models (HMMs)~\cite{knill1997hidden} and Gaussian Mixture Models (GMMs)~\cite{reynolds2009gaussian}, which were used to generate sequential data. However, significant advancements in generative models occurred with the rise of deep learning. In natural language processing (NLP), traditional methods for sentence generation involved learning word distributions using N-gram language models~\cite{bengio2000neural} and then searching for the best sequence. To handle longer sentences, recurrent neural networks (RNNs)~\cite{mikolov2010recurrent} were introduced for language modeling tasks, allowing for the modeling of relatively long dependencies, a capability enhanced by Long Short-Term Memory (LSTM) and Gated Recurrent Units (GRU), which use gating mechanisms to control memory during training. These methods can effectively attend to approximately 200 tokens in a sample manner~\cite{khandelwal2018sharp}, marking a substantial improvement over N-gram models. In computer vision (CV), Generative Adversarial Networks (GANs)~\cite{goodfellow2020generative} have achieved remarkable results across various applications. Additionally, Variational Autoencoders (VAEs)~\cite{kingma2013auto} and diffusion models~\cite{song2019generative} have been developed to provide more fine-grained control over the image generation process, enabling the creation of high-quality images.

\subsection{Diffusion Models}

Diffusion models are powerful tools for generating synthetic data. The Denoising Diffusion Probabilistic Model (DDPM) is a type of latent variable model inspired by nonequilibrium thermodynamics, using a Gaussian distribution for data generation~\cite{nichol2021improved}. These models are not only simple to define but also efficient to train, and they can be integrated with non-autoregressive text generation methods to improve text generation quality~\cite{li2023diffusion}. Song~\textit{et al.}~\cite{song2020score} introduced a stochastic differential equation (SDE) that gradually transforms a complex data distribution into a known prior distribution by adding noise, and a reverse-time SDE that reconstructs the data distribution from the prior by gradually removing the noise. The reverse-time SDE relies solely on the time-dependent gradient field of the perturbed data distribution. Vahdat~\textit{et al.}~\cite{vahdat2021score} proposed the Latent Score-based Generative Model (LSGM), a new method that trains Score-based Generative Models (SGMs) in a latent space within the framework of variational autoencoders for image generation.

\subsection{Reweighting Samples for AI Fairness}

AI fairness has emerged as one of the most critical challenges of the decade~\cite{shaham2023holistic}. Although machine learning models are designed to intelligently avoid errors and biases in decision-making, they can sometimes unintentionally perpetuate bias and discrimination within society. Concerns have been raised about various forms of unfairness in ML, including racial biases in criminal justice, disparities in employment, and biases in loan approvals~\cite{angwin2022machine}. The entire lifecycle of an ML model, from input data through modeling, evaluation, and feedback, is vulnerable to both external and inherent biases, which can lead to unjust outcomes. Techniques to mitigate bias in ML models are generally divided into three categories: pre-processing, in-processing, and post-processing~\cite{caton2020fairness}. Pre-processing recognizes that data itself can introduce bias, with distributions of sensitive or protected variables often being discriminatory or imbalanced. For example, Blow~\textit{et al.}~\cite{blow2024comprehensive} conducted a systematic study of reweighting samples for traditional ML models, using five models for binary classification on datasets such as Adult Income and COMPAS, and incorporating various protected attributes. Notably, the study leveraged AI Fairness 360 (AIF360), a comprehensive open-source library designed to identify and mitigate bias in machine learning models throughout the AI application lifecycle.

\section{Conclusion and Future Work}
\label{sec7}

Understanding the impact of generative modeling is crucial to preventing unintended bias when augmenting training data. This study explores data augmentation via diffusion models, aiming to reduce bias and improve overall performance. It involved evaluating model performance with the generated data added in various increments to the original dataset, and comparing the results to the original outputs using metrics including balanced accuracy and fairness metrics. Experimental results indicated the effectiveness of synthetic data generated by diffusion models for data augmentation. Future work will build on this exploration by incorporating additional datasets and comparing the effects of varying data increments. Additionally, different tools from AI Fairness 360 (AIF360) will be tested to further mitigate bias.

\section*{Acknowledgment}
\label{acknowledgement}
This research work is supported by NSF  under award number 2323419 and by the Army Research Office under Cooperative Agreement Number W911NF-24-2-0133. The views and conclusions contained in this document are those of the authors and should not be interpreted as representing the official policies, either expressed or implied, of the NSF or the Army Research Office or the U.S. Government. The U.S. Government is authorized to reproduce and distribute reprints for Government purposes notwithstanding any copyright notation herein.

\ifCLASSOPTIONcaptionsoff
  \newpage
\fi


\bibliographystyle{IEEEtran}
\bibliography{Reference}

\end{document}